\documentclass[10pt,twocolumn,letterpaper]{article}

\usepackage[pagenumbers]{cvpr} %

\usepackage{graphicx}
\usepackage{amsmath}
\usepackage{amssymb}
\usepackage{booktabs}

\usepackage{xcolor}         %
\usepackage{pifont}
\newcommand{\cmark}{\ding{51}}
\newcommand{\xmark}{\ding{55}}
\definecolor{myred}{RGB}{220,50,47} %
\definecolor{mygreen}{RGB}{133,153,0}
\definecolor{commentcolor}{RGB}{133,153,0}
\newcommand{\greencmark}{\textcolor{mygreen}{\cmark}}
\newcommand{\redxmark}{\textcolor{myred}{\xmark}}

\newcommand*{\affmark}[1][*]{\textsuperscript{#1}}
\definecolor{urlcolor}{rgb}{0.93,0.01,0.55}

\usepackage[pagebackref,breaklinks,colorlinks]{hyperref}
\usepackage{enumitem}
\usepackage{color,colortbl}
\usepackage{tocloft}
\usepackage{titletoc}

\usepackage[capitalize]{cleveref}
\crefname{section}{Sec.}{Secs.}
\Crefname{section}{Section}{Sections}
\Crefname{table}{Table}{Tables}
\crefname{table}{Tab.}{Tabs.}

\def\cvprPaperID{3956} %
\def\confName{CVPR}
\def\confYear{2023}

\begin{document}

\title{SINE: \underline{SIN}gle Image \underline{E}diting with Text-to-Image Diffusion Models}

\author{
Zhixing Zhang\affmark[1]\quad
Ligong Han\affmark[1]\thanks{Corresponding author: \texttt{ligong.han@rutgers.edu}}\quad
Arnab Ghosh\affmark[2]\quad
Dimitris Metaxas\affmark[1]\quad
Jian Ren\affmark[2]\\
\affmark[1]Rutgers University\quad\quad
\affmark[2]Snap Inc.
}

\maketitle

\begin{abstract}
Recent works on diffusion models have demonstrated a strong capability for conditioning image generation, e.g., text-guided image synthesis.
Such success inspires many efforts trying to use large-scale pre-trained diffusion models for tackling a challenging problem--real image editing.
Works conducted in this area learn a unique textual token corresponding to several images containing the same object. However, under many circumstances, only one image is available, such as the painting of the Girl with a Pearl Earring. Using existing works on fine-tuning the pre-trained diffusion models with a single image causes severe overfitting issues. The information leakage from the pre-trained diffusion models makes editing can not keep the same content as the given image while creating new features depicted by the language guidance. 
This work aims to address the problem of single-image editing. We propose a novel model-based guidance built upon the classifier-free guidance so that the knowledge from the model trained on a single image can be distilled into the pre-trained diffusion model, enabling content creation even with one given image.
Additionally, we propose a patch-based fine-tuning that can effectively help the model generate images of arbitrary resolution. We provide extensive experiments to validate the design choices of our approach and show promising editing capabilities, including changing style, content addition, and object manipulation\footnote{ 
{More visual examples on our~\href{https://zhang-zx.github.io/SINE/}{\color{urlcolor}{Webpage}}.}}.

\end{abstract}
\section{Introduction}

Automatic real image editing is an exciting direction, enabling content generation and creation with minimal effort. Although many works have been conducted in this area, achieving high-fidelity semantic manipulation on an image is still a challenging problem for the generative models, considering the target image might be out of the training data distribution~\cite{kim2021exploiting, gal2022stylegan, zhu2017unpaired, zhu2020domain, meng2021sdedit, bar2022text2live, kim2022diffusionclip}.
The recently introduced large-scale text-to-image models, \emph{e.g.}, DALL·E 2~\cite{ramesh2022hierarchical}, Imagen~\cite{saharia2022photorealistic}, Parti~\cite{yu2022scaling}, and StableDiffusion~\cite{rombach2022high}, can perform high-quality and diverse image generation with natural language guidance.
The success of these works has inspired many subsequent efforts to leverage the pre-trained large-scale models for real image editing~\cite{ruiz2022dreambooth, hertz2022prompt, gal2022image}. They show that, with properly designed prompts and a limited number of fine-tuning steps, the text-to-image models can manipulate a given subject with text guidance. 
\begin{figure}[t!]
\centering
\includegraphics[width=1\linewidth]{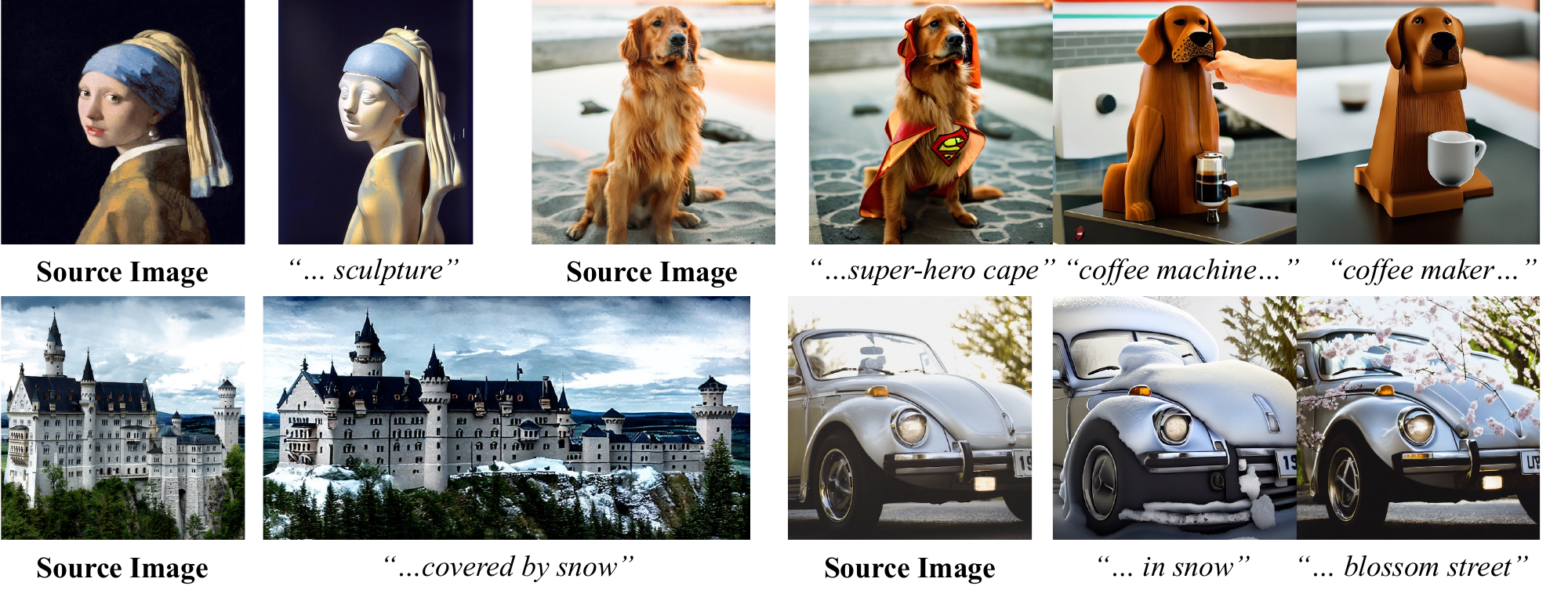}
\caption{With only \emph{one} real image, \emph{i.e.}, Source Image, our method is able to manipulate and generate the content in various ways, such as changing style, adding context, modifying the object, and enlarging the resolution, through guidance from the text prompt.}
\label{fig:teaser}
\end{figure}

On the downside, the recent text-guided editing works that build upon the diffusion models suffer several limitations. 
First, the fine-tuning process might lead to the pre-trained large-scale model overfit on the real image, which degrades the synthesized images' quality when editing. To tackle these issues, methods like using multiple images with the same content and applying regularization terms on the same object have been introduced~\cite{ruiz2022dreambooth,bar2022text2live}. However, querying multiple images with identical content or object might not be an available choice; for instance, there is only one painting for \emph{Girl with a Pearl Earring}.
Directly editing the single image brings information leakage from the pre-trained large-scale models, generating images with different content (examples in Fig.~\ref{fig:Comparison}); therefore, the application scenarios of these methods are greatly constrained. Second, these works lack a reasonable understanding of the object geometry for the edited image. Thus, generating images with different spatial size as the training data cause undesired artifacts, \emph{e.g.}, repeated objects, and incorrectly modified geometry (examples in Fig.~\ref{fig:Comparison}). Such drawbacks restrict applying these methods for generating images with an arbitrary resolution, \emph{e.g.}, synthesizing high-resolution images from a single photo of a castle (as in Fig.~\ref{fig:resolution_demo}), again limiting the usage of these methods.

In this work, we present SINE, a framework utilizing pre-trained text-to-image diffusion models for \textbf{SIN}gle image \textbf{E}diting and content manipulation.
We build our approach based upon existing text-guided image generation approaches~\cite{ruiz2022dreambooth,gal2022image} and propose the following novel techniques to solve overfitting issues on content and geometry, and language drift~\cite{lee2020countering, lu2020countering}:
\begin{itemize}[leftmargin=1em]
    \item First, by appropriately modifying the classifier-free guidance~\cite{ho2021classifier}, we introduce model-based classifier-free guidance that utilizes the \emph{diffusion model} to provide the score guidance for content and structure. Taking advantage of the step-by-step sampling process used in diffusion models, we use the model fine-tuned on a single image to plant a content ``seed'' at the early stage of the denoising process and allow the pre-trained large-scale text-to-image model to edit creatively conditioned with the language guidance at a later stage. 
    \item Second, to decouple the correlation between pixel position and content, we propose a patch-based fine-tuning strategy, enabling generation on arbitrary resolution.
\end{itemize}

With a text descriptor describing the content that is aimed to be manipulated and language guidance depicting the desired output, our approach can edit the \emph{single} unique image to the \emph{targeted domain} with details preserved in \emph{arbitrary} resolution.
The output image keeps the structure and background intact while having features well-aligned with the target language guidance.
As shown in Fig.~\ref{fig:teaser}, trained on a painting \textit{Girl with a Pearl Earring} with resolution as $512\times512$, we can sample an image of a sculpture of the girl at the resolution of $640\times512$ with the identity features preserved. 
Moreover, our method can successfully handle various edits such as style transfer, content addition, and object manipulation (more examples in Fig.~\ref{fig:editing results}). We hope our method can further boost creative content creation by opening the door to editing arbitrary images.

\section{Related Work}
Text-guided image synthesis has drawn considerable attention in the generative model context~\cite{zhu2019dm, tao2020df, xu2018attngan, zhang2021cross, ye2021improving, ramesh2021zero, yu2022scaling, oh2001image, abdal2021styleflow, harkonen2020ganspace, patashnik2021styleclip,abdal2022clip2stylegan}.
The recent development of diffusion models~\cite{ho2020denoising, song2021denoising, sohl2015deep, song2019generative} introduced new solutions to this problem and produced impressive results~\cite{ramesh2022hierarchical, saharia2022photorealistic, nichol2021glide, rombach2022high}.
With the significant improvement of these models, rather than training a large-scale text-to-image model from scratch, a leading line of works focuses on taking advantage of the existing pre-trained model and manipulating images according to given natural language guidance~\cite{hertz2022prompt, gal2022image, ruiz2022dreambooth, kawar2022imagic, liu2021more, avrahami2022blended}.
In these works, studies explore text-based interfaces for image editing~\cite{patashnik2021styleclip, abdal2022clip2stylegan, avrahami2022blended}, style transfer~\cite{kwon2022diffusion, liu2022name}, and generator domain adaption~\cite{gal2022stylegan, kim2022diffusionclip}.

The development of the diffusion model provides a giant and flexible design space for this task.
Many works utilize pre-trained diffusion models as generative priors and are training-free. ILVR~\cite{choi2021ilvr} guides the denoising process by replacing the low-frequency part of the sample with that of the target reference image. SDEdit~\cite{meng2021sdedit} applies the diffusion process first on an image or a user-created semantic map and then conducts the denoising procedure conditioned with the desired output. Blended diffusion~\cite{avrahami2022blended} performs language-guided inpainting with a given mask.

Another line of research showed great potential and semantic editing ability of fine-tuning. DiffusionCLIP~\cite{kim2022diffusionclip} leverages the CLIP~\cite{radford2021learning} model to provide gradients for image manipulation and delivers impressive results on style transfer. Textual-Inversion~\cite{gal2022image} and DreamBooth~\cite{ruiz2022dreambooth} fine-tune the text embedding or the full diffusion model using a few personalized images (typically $3\sim5$) to synthesize images of the same object in a novel context. These methods, however, either drastically change the layout of the original image when dealing with a single image or can not fully leverage the generalization ability of the pre-trained model for editing due to overfitting or language drift. Notably, Prompt-to-Prompt~\cite{hertz2022prompt} controls the editing of synthesized images by manipulating the cross-attention maps; however, its editing ability is limited when applied to real images.

This work introduces a solution to achieve image fidelity and text alignment simultaneously.
Our method can perform high-quality semantic editing globally and locally on one single image.
On the other hand, previous works lack an understanding of the object geometry of the edited image.
When editing the image at an arbitrary resolution, the artifacts in the results will be obvious.
Prior works have investigated generating images at arbitrary resolution using positional encoding as inductive bias~\cite{xu2021positional,xu2021positional,ryu2022pyramidal} so that the correlation between content and position can be eliminated.
Anyres-GAN~\cite{chai2022any} adopt a patch training mechanism to leverage high-resolution data to help the generation of images in the low-resolution domain.
We propose a patch-based fine-tuning method to achieve arbitrary resolution editing.

\section{Methods}\label{sec: methods}

For \emph{one} arbitrary in-the-wild image, our goal is to edit the image via language while preserving the maximal amount of details from the original image.
To do so, we leverage the generalization ability of pre-trained large-scale text-to-image models~\cite{rombach2022high}.
An intuitive approach is to fine-tune the diffusion models with the single image and text description, similar to DreamBooth~\cite{ruiz2022dreambooth}.
Ideally, it should provide a model that can reconstruct the input image using the given text descriptor and synthesize new images when given other language guidance.
Unfortunately, we find the model can easily overfit the single trained image and its corresponding text description. Thus, although the fine-tuned model can still reconstruct the input image perfectly, it can no longer synthesize diverse images according to the given language guidance (as shown in Fig.~\ref{fig:Comparison}). 
Moreover, it struggles to generate arbitrary resolution images due to the lack of positional information (as in Fig.~\ref{fig:resolution_demo}).

\begin{figure}
  \centering
  \includegraphics[width=1\linewidth]{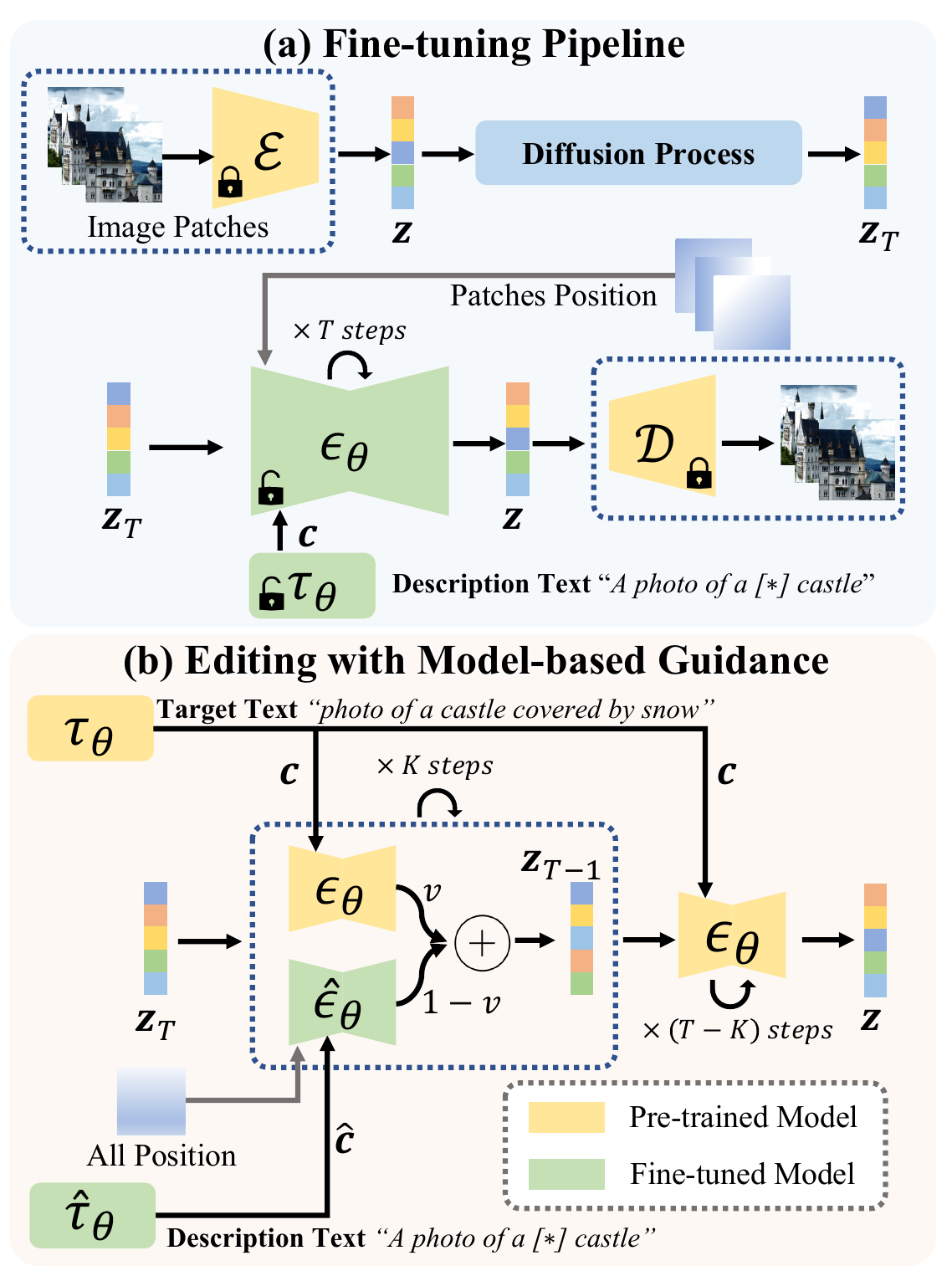}
  \caption{\textbf{Overview of our method.} 
  (a) Given a source image, we first randomly crop it into patches and get the corresponding latent code $\mathbf{z}$ with the pre-trained encoder.
   At fine-tune time, the denoising model, $\boldsymbol{\epsilon}_\theta$, takes three inputs: noisy latent $\mathbf{z}_T$, language condition $\mathbf{c}$, and positional embedding for the area where the noisy latent is obtained. 
  (b) During sampling, we give additional language guidance about the target domain to edit the image. 
  Also, we sample a noisy latent code $\mathbf{z}_T$ with the dimension corresponding to the desired output resolution.
  Language conditioning for $\boldsymbol{\epsilon}_\theta$ and $\mathbf{c}$ are given by pre-trained language encoder $\boldsymbol{\tau}_\theta$ with the target language guidance.
  While for the fine-tuned diffusion model, $\boldsymbol{\hat{\epsilon}}_\theta$, in addition to the language conditioning $\mathbf{\hat{c}}$, we also input the positional embedding for the whole image.
  We employ a linear combination between the score calculated by each model for the first $K$ steps and inference only on pre-trained $\boldsymbol{\epsilon}_\theta$ after.
  }
  \label{fig:overview}
\end{figure}

To solve the above issues, we propose a test-time model-based classifier-free guidance and a patch-based fine-tuning technique.
An overview of our method is illustrated in Fig.~\ref{fig:overview}.
In the following sections, we review the backbone model used in our approach (Sec.~\ref{sec: methods: diffusion models}). Then, we describe how to overcome the overfitting problem with model-based guidance (Sec.~\ref{sec: methods: guidance}). Lastly, we present how to address the problem of limited resolution generation (Sec.~\ref{sec: methods: finetune}).

\subsection{Language-Guided Diffusion Models}\label{sec: methods: diffusion models}
We use the latent diffusion models (LDMs)~\cite{rombach2022high} trained on a large-scale dataset as our base model and implement the proposed approaches by fine-tuning the pre-trained model.
LDMs is a class of Denoising Diffusion Probabilistic Models (DDPMs)~\cite{ho2020denoising} that contains an auto-encoder trained on images, and a diffusion model learned on the latent space constructed by the auto-encoder.
The encoder $\mathcal{E}$ encodes a given image $\mathcal{I} \in \mathbb{R}^{H \times W \times 3}$ to a latent representation $\mathbf{z}$, such that $\mathbf{z}=\mathcal{E}(\mathcal{I})$. The decoder $\mathcal{D}$ reconstructs the estimated image $\tilde{\mathcal{I}}$ from the latent, such that $\tilde{\mathcal{I}}=\mathcal{D}(\mathbf{z})$ and $\tilde{\mathcal{I}} \approx \mathcal{I}$.
The diffusion model is trained to produce latent codes within the pre-trained latent space.
The most intriguing property of LDMs is that the diffusion model can be conditioned on class labels, images, and text prompt.
The conditional LDM is learned as follows:
\begin{equation}
\small
L_{L D M}:=\mathbb{E}_{\mathcal{E}(\mathcal{I}), y, \boldsymbol{\epsilon} \sim \mathcal{N}(0,1), t}\left[\left\|\boldsymbol{\epsilon}-\boldsymbol{\epsilon}_\theta\left(\mathbf{z}_t, t, \boldsymbol{\tau}_\theta(y)\right)\right\|_2^2\right],
\label{equ: ldm}
\end{equation}
where $t$ is the time step, $\mathbf{z}_t$ is the latent noised to time $t$, $\boldsymbol{\epsilon}$ is the unscaled noise sample, $\boldsymbol{\epsilon}_\theta$ is the denoising model, $y$ is the conditioning input, and $\boldsymbol{\tau}_\theta$ maps $y$ to a conditioning vector.
During training time, $\boldsymbol{\epsilon}_\theta$ and $\boldsymbol{\tau}_\theta$ are jointly optimized.
A random noise tensor is sampled and denoised at inference time based on the conditioning input, \eg, text prompt, to produce a new latent.
Inspired by DreamBooth~\cite{ruiz2022dreambooth}, we construct the text prompt for fine-tuning a single image as ``a photo/painting of a [$\ast$] [class noun]'', where ``[$\ast$]'' is a unique identifier and ``[class noun]'' is a coarse class descriptor (\eg, ``castle'', ``lake'', ``car'', \etc).

\subsection{Model-Based Classifier-Free Guidance}
\label{sec: methods: guidance}
With the above-presented LDMs, we introduce our approach, inspired by classifier-free guidance, to overcome overfitting when fine-tuning LDMs with one image.

\noindent\textbf{Classifier-free guidance}~\cite{ho2021classifier} is a technique widely adopted
by prior text-to-image diffusion models~\cite{rombach2022high, saharia2022photorealistic}.
A single diffusion model is trained using conditional and unconditional objectives by randomly dropping the condition during training.
When sampling, a linear combination of the conditional and unconditional score estimation is used:
\begin{equation}
\small
\tilde{\boldsymbol{\epsilon}}_\theta\left(\mathbf{z}_t, \mathbf{c}\right)=w \boldsymbol{\epsilon}_\theta\left(\mathbf{z}_t, \mathbf{c}\right)+(1-w) \boldsymbol{\epsilon}_\theta\left(\mathbf{z}_t\right),
\label{equ: classifier-free}
\end{equation}
where $\boldsymbol{\epsilon}_\theta\left(\mathbf{z}_t, \mathbf{c}\right)$ and $\boldsymbol{\epsilon}_\theta\left(\mathbf{z}_t\right)$ are the conditional and unconditional $\boldsymbol{\epsilon}$-predictions, $\mathbf{c}$ is the conditioning vector generated by $\tau_\theta$, and $w$ is the weight for the guidance.
The predication is performed using the Tweedie's formula~\cite{chung2022diffusion}, namely, $\left(\mathbf{z}_t-\sqrt{1-\bar{\alpha}_t} \tilde{\boldsymbol{\epsilon}}_\theta\right) / \sqrt{\bar{\alpha}_t}$, where $\bar{\alpha}_t$ is a function of $t$ that affects the sampling quality.

Since we only have one image as the training data, \emph{e.g.}, painting of \emph{Mona Lisa}, and one corresponding text descriptor of that image, the diffusion model suffers from overfitting, and severe language drifts after fine-tuning~\cite{ruiz2022dreambooth}. As a result, the fine-tuned model fails to synthesize images containing features from other language guidance. The overfitting issue might be due to only one repeated prompt used during fine-tuning, making other text prompts no longer accurate enough to control editing (see examples in Fig.~\ref{fig: guidance_ablation}).

\noindent\textbf{Model-based classifier-free guidance.}
Existing ``personalized" text-guided real image editing works only use \emph{one} fine-tuned model for image generation and editing~\cite{ruiz2022dreambooth, hertz2022prompt, gal2022image}, ignoring the capacity of pre-trained large-scale text-to-image models. Instead, to alleviate the overfitting of the fine-tuned model, we leverage the \emph{pre-trained text-to-image model for image generation} with the provided language guidance and use the fine-tuned model to provide content features in a fashion of combining scores from the two models, similar to classifier-free guidance.

Specifically, let $\boldsymbol{\hat{\epsilon}}_\theta$ denote the fine-tuned denoising model, and $\boldsymbol{\epsilon}_\theta$ denote the pre-trained text-to-image model.
During sampling, at specified steps, we use our fine-tuned model to guide the pre-trained one by using a linear combination of the scores from each model.
Thus, the score estimation in Eqn.~\ref{equ: classifier-free} becomes:
\begin{equation}
\small
\begin{aligned}
\tilde{\boldsymbol{\epsilon}}_\theta\left(\mathbf{z}_t, \mathbf{c}\right)=& w\left(v \boldsymbol{\epsilon}_\theta\left(\mathbf{z}_t, \mathbf{c}\right)+(1-v) \hat{\boldsymbol{\epsilon}}_\theta\left(\mathbf{z}_t, \mathbf{\hat{c}}\right)\right) \\
&+(1-w) \boldsymbol{\epsilon}_\theta\left(\mathbf{z}_t\right),
\end{aligned}
\label{equ: modified classifier-free}
\end{equation}
where $v$ stands for the model guidance weight, $\mathbf{\hat{c}}$ is the language guidance token obtained from the fine-tuned diffusion model with the text prompt used during fine-tuning, and $\mathbf{c}$ is the target language conditioning obtained from the target prompt.

To prevent artifacts from the over-fitted model and maintain the fidelity of the generated image, we propose to sample using Eqn.~\ref{equ: modified classifier-free} with $t>K$ and sample using Eqn.~\ref{equ: classifier-free} for $t \leq K$.
From $K$ to 0, the denoising process only depends on the pre-trained model.
Following this approach, we can fully leverage the generalization ability of the pre-trained model (examples in Fig.~\ref{fig: guidance_ablation}).
{Also note that this method could be generalized to include multiple prompts or even multiple modalities.}

\subsection{Patch-Based Fine-Tuning}\label{sec: methods: finetune}
With model-based classifier-free guidance, we are able to edit and manipulate a single image with given language guidance. Here, we further show how to improve the fine-tuning process for a single training image so that the fine-tuned model can better understand the content and geometry of the image. Thus, it can provide better content guidance for the large-scale text-to-image model during the sampling time and unleash the potential for generating arbitrary-resolution images~\cite{esser2021taming}.

\noindent\textbf{Limited-resolution generation.} We first review the limitations of the current fine-tuning process. Given an input image $\mathcal{I}$ with resolution as $H\times W$, we can obtain a downsampled latent code $\mathbf{z}$ from the pre-trained encoder.
Since the text-to-image diffusion model is pre-trained at a fixed resolution, \emph{i.e.}, $p \times p$, we need to resize the input image to a corresponding resolution $sp \times sp$, where $s$ represents the scaling factor of the encoder, to match the resolution for reducing the fine-tuning cost.
In essence, prior knowledge of the correlation between the position and content information is learned by the diffusion model.
Thus, when sampling from a higher-resolution noise tensor, the generated latent code leads to artifacts like duplicates or position shifting (visual examples in Fig.~\ref{fig:Comparison}).
To tackle such drawbacks, we propose a simple yet effective fine-tuning method.
\begin{figure*}
\setlength{\linewidth}{\textwidth}
\setlength{\hsize}{\textwidth}
\centering
\includegraphics[width=1\linewidth]{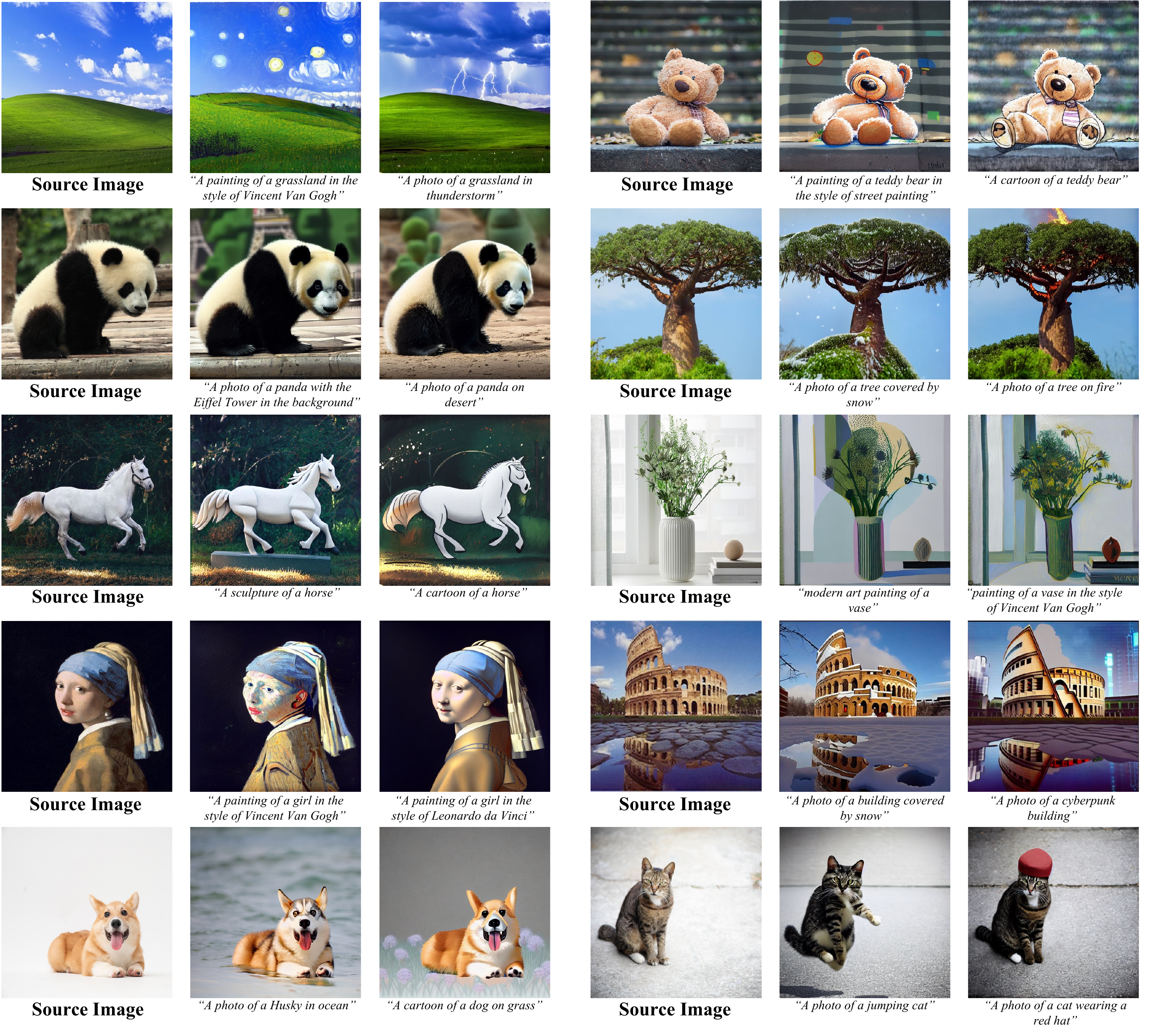}
\caption{\textbf{Editing on single source image from various domains.} We employ our method on various images and edit them with two target prompts at $512\times512$ resolution. We show the wide range of edits our approach can be used, including but not limited to style transfer, content add-on, posture change, breed change, \etc}
\label{fig:editing results}

\end{figure*}

\noindent\textbf{Patch-based fine-tuning.}
Inspired by Chai \emph{et al.}~\cite{chai2022any}, we treat our single training image as a function on coordinate for each pixel, bounded in $[0, H]\times [0, W]$.
The diffusion model still generates latent code at the fixed resolution $p \times p$, but each latent code corresponds to a sub-area in the image.
We denote the sub-area as $\mathbf{v}=[h_1, w_1, h_2, w_2]$, where $(h_1, w_1) \in [0, H]\times [0, W]$ and $(h_2, w_2) \in (h_1, H]\times (w_1, W]$ indicate the top-left and bottom-right coordinates of the area, respectively.
During fine-tuning, we sample patches from the image with different $\mathbf{v}$ and resize the patches to resolution $sp$.
We denote the resulted patch as $\mathcal{I}(F(\mathbf{v})) \in \mathbb{R}^{sp\times sp \times 3}$, where $F$ is the normalization and Fourier embedding~\cite{karras2021alias} of the specific area.
The encoded latent code of the patch is $\mathbf{z}_{\mathbf{v}} = \mathcal{E}(\mathcal{I}(F(\mathbf{v})))$.
Our model uses the normalized Fourier embedding as an input to make the model learn the position-content correlation.
Formally, our diffusion model is defined as $\hat{\boldsymbol{\epsilon}}_\theta\left(\mathbf{z}_t, t, \boldsymbol{\tau}_\theta(y), F(\mathbf{v})\right)$.

After fine-tuning, the model can generate latent code at different resolutions by giving the positional information directly to the model.
The arbitrary resolution image editing is conducted by feeding two inputs to the model: the positional embedding of the whole image; and a randomly sampled noisy latent with the dimension corresponding to the resolution we want.
When sampling in an arbitrary resolution, the model can still keep the structure of the original image intact (examples in Fig.~\ref{fig:resolution_demo}). 
It is worth noting that with or without the correct position encoding, our framework still naturally permits retargeting, \ie, maintaining the aspect ratio of salient objects, like SinGAN\cite{shaham2019singan}, InGAN\cite{shocher2019ingan}, and Drop-the-GAN\cite{granot2022drop}.

\section{Experiments}

\noindent\textbf{Implementation Details}
While our method can be generally applied to different frameworks, we implement it based on the recently released text-to-image LDM, Stable Diffusion~\cite{rombach2022high}.
The pre-trained model was pre-trained on $512\times512$ images from LAION dataset~\cite{schuhmann2021laion}.
The spatial size of the latent code from the pre-trained model is $64\times 64$.

For patch-based fine-tuning, we randomly crop images to patches with height and width uniformly in the range of $[0.1H, H]\times[0.1W, W]$ and resize them to $512\times512$.
Experiments are conducted using $1\times \text{RTX 8000}$ GPU with a batch size of $1$.
The base learning rate is set to $1\times10^{-6}$.
The number of time steps for the diffusion model, $T$, is $1000$.
Experiments without and with patch-based fine-tuning are created  after $800$ and $10,000$ optimization steps, respectively.
Unless otherwise noted, we adopt other hyperparameter choices from Stable Diffusion~\cite{rombach2022high}, and the results are generated with image resolution $512\times512$ and with latent dimension $64\times64$.
For sampling parameters, we choose $K=400$ and $v=0.7$.

\subsection{Qualitative Evaluation}
\label{sec: exp: Qualitative}

To better understand various approaches, we collect images from a wide range of domains, \emph{i.e.}, free-to-use high-resolution images from Flickr\footnote{https://www.flickr.com/} and Unsplash\footnote{https://unsplash.com/}.
During fine-tuning, we apply a coarse class descriptor to the content we want to preserve, \eg, dog, cat, castle, \etc.
After optimization, we edit each image with diverse editing prompts.
We randomly generate $4$ edit results for each image and editing prompt and choose the best one (such a process is also applied to other comparison methods).
Our work shows impressive editing ability when applied to various images with different language guidance.

\begin{figure}
\centering
\includegraphics[width=\linewidth]{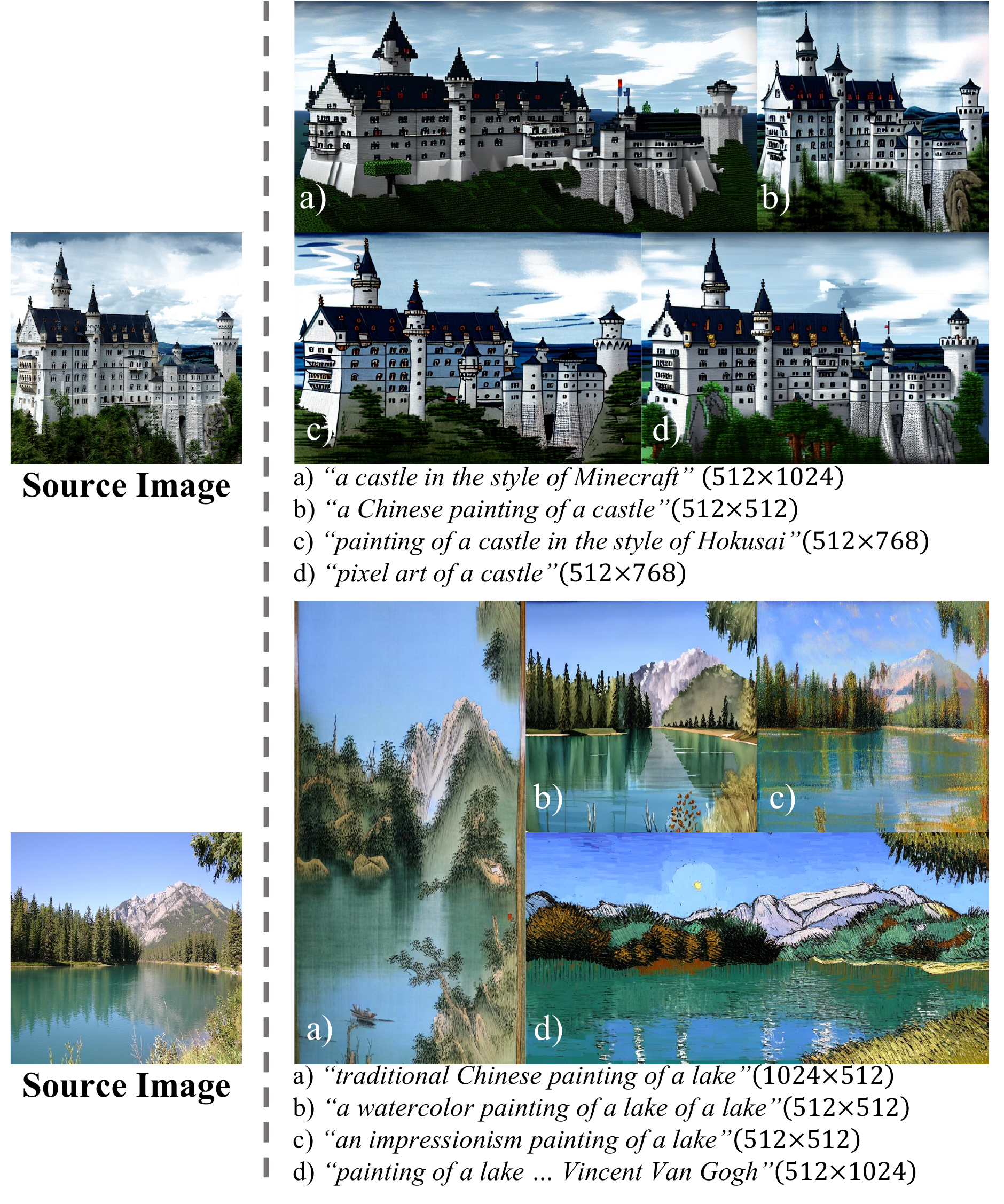}
\caption{\textbf{Arbitrary resolution editing.} Our method achieves higher-resolution image editing without artifacts like duplicates, even on ones that change the height-width ratio drastically.}
\label{fig:resolution_demo}
\end{figure}

As presented in Fig.~\ref{fig:editing results}, using the model-based classifier-free guidance (Sec.~\ref{sec: methods: guidance}) enables us to apply various editing via text prompts on the \emph{single} real images.
Each image has two text prompts describing different features we want to edit, \eg, image style, background content, the texture of the content, \etc.
Our method can edit the related features while keeping the content intact.
We further show our editing results on arbitrary resolution generation in Fig.~\ref{fig:resolution_demo}.
For each source image, we edit it with different prompts at various resolutions.
As can be seen, our patch-based fine-tuning schedule (Sec.~\ref{sec: methods: finetune}) successfully preserves the original portion and geometry features of the single source image, even on highly challenging resolution such as $512\times1024$.

\begin{figure*}
\setlength{\linewidth}{\textwidth}
\setlength{\hsize}{\textwidth}
\centering
\includegraphics[width=\textwidth]{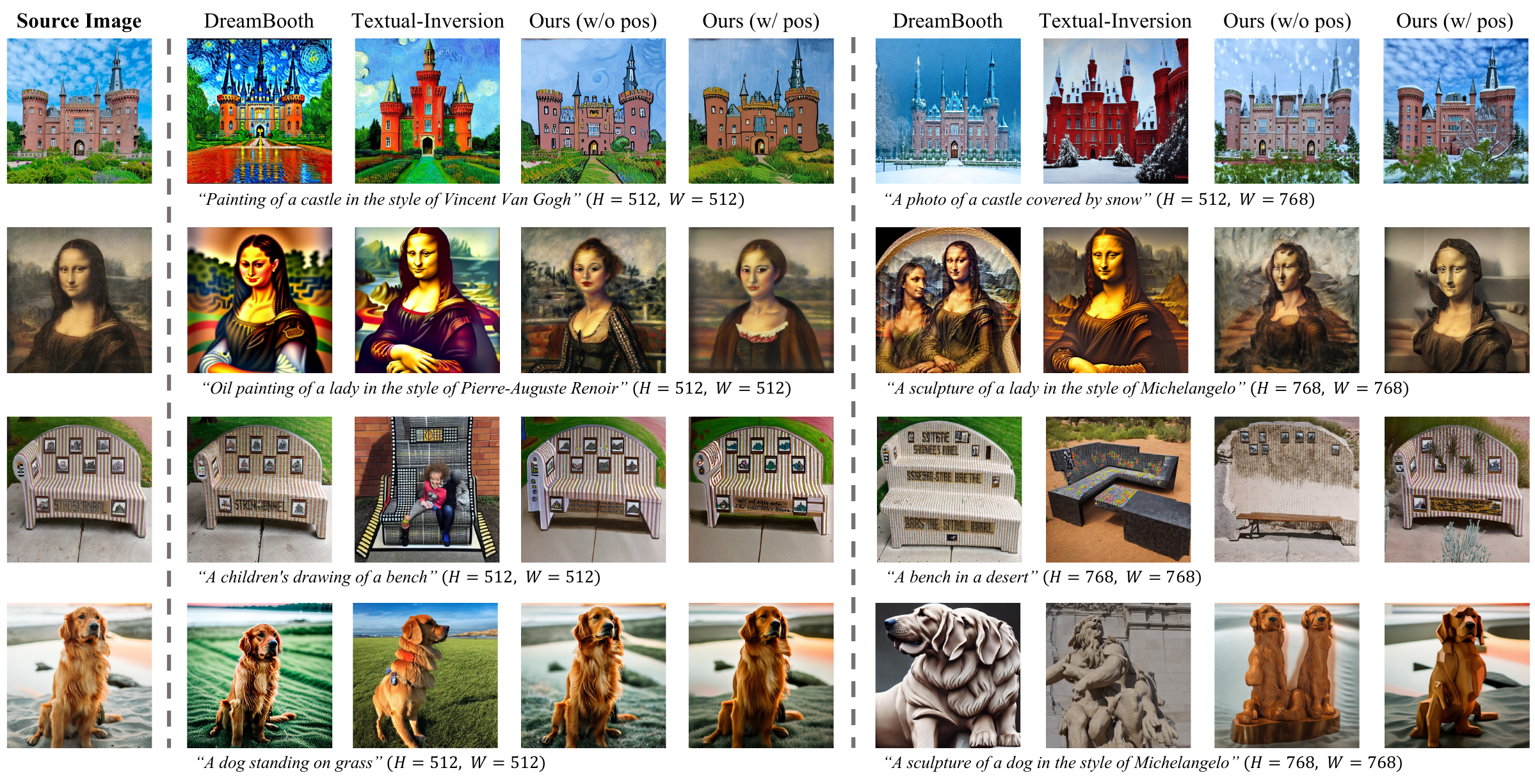}
\caption{\textbf{Comparisons of various methods.} We compare our method to DreamBooth~\cite{ruiz2022dreambooth} and Textual-Inversion~\cite{gal2022image}. On the left part of the figure, we edit at the resolution \emph{same} as training time. On the right part, we edit the source image at a \emph{higher} resolution. Our work successfully edits the image as required while preserving the details of the source images. We also compare our method without and with the patch-based fine-tuning mechanism (w/o pos \emph{vs.} w/ pos). When editing at a fixed resolution, two settings perform equally, while at a higher resolution, the patch-based fine-tuning method successfully prevents artifacts. }
\label{fig:Comparison}
\end{figure*}

\subsection{Comparisons}
\label{sec: exp: comp}

We compare our method to concurrent leading techniques, Textual-Inversion~\cite{gal2022image} and DreamBooth~\cite{ruiz2022dreambooth}, that can be used for single-image editing.
Considering no official implementation has been released for DreamBooth, we adopt an unofficial but well-adopted and highly competitive implementation based on Stable Diffusion\cite{drembooth_github,rombach2022high}.
We compare these techniques strictly according to the detailed guidance provided with the implementations.

Fig.~\ref{fig:Comparison} shows the comparison results.
As can be noticed, our method maintains the fidelity of the images while applying changes as desired. Furthermore, our approach has high authenticity and structural integrity even for higher-resolution editing.
For example, in the last row of Fig.~\ref{fig:Comparison}, when the target prompt is ``... \emph{standing on grass}", our method generates results by modifying the texture of the land on which the dog stands with other features intact.
However, other methods result in a dramatic change in the structure of the whole image.
Moreover, in the second row, when modifying the painting \emph{Mona Lisa}, both DreamBooth~\cite{ruiz2022dreambooth} and Textual-Inversion~\cite{gal2022image} fail to edit the image.
Our work also shows clear advantages over the approaches on training-free editings, such as ILVR~\cite{choi2021ilvr}, SDEdit~\cite{meng2021sdedit}, and Prompt-to-prompt~\cite{hertz2022prompt}, with the qualitative comparisons presented in the Appendix.

\subsection{Ablation Analysis}
\label{sec: exp: ablation}

\noindent \textbf{Patch-based fine-tuning.}
In Fig.~\ref{fig:Comparison}, we show the results of editing images in higher resolution when fine-tuned without or with the proposed patch-based fine-tuning technique (w/o pos \vs w/ pos).
When sampling at a higher resolution, as in the right part of Fig.~\ref{fig:Comparison}, the denoising model fine-tuned without the patch-based training mechanism performs poorly.
In the first row, the castle towers get duplicated to meet the resolution, and in the third row, the bench gets stretched disproportionately.
In essence, our patch-based fine-tuning technique enables the diffusion model to leverage the super-resolution ability of the decoder and edit images at arbitrary resolution during testing time.

\begin{figure}[ht]
    \centering
    \includegraphics[width=1\linewidth]{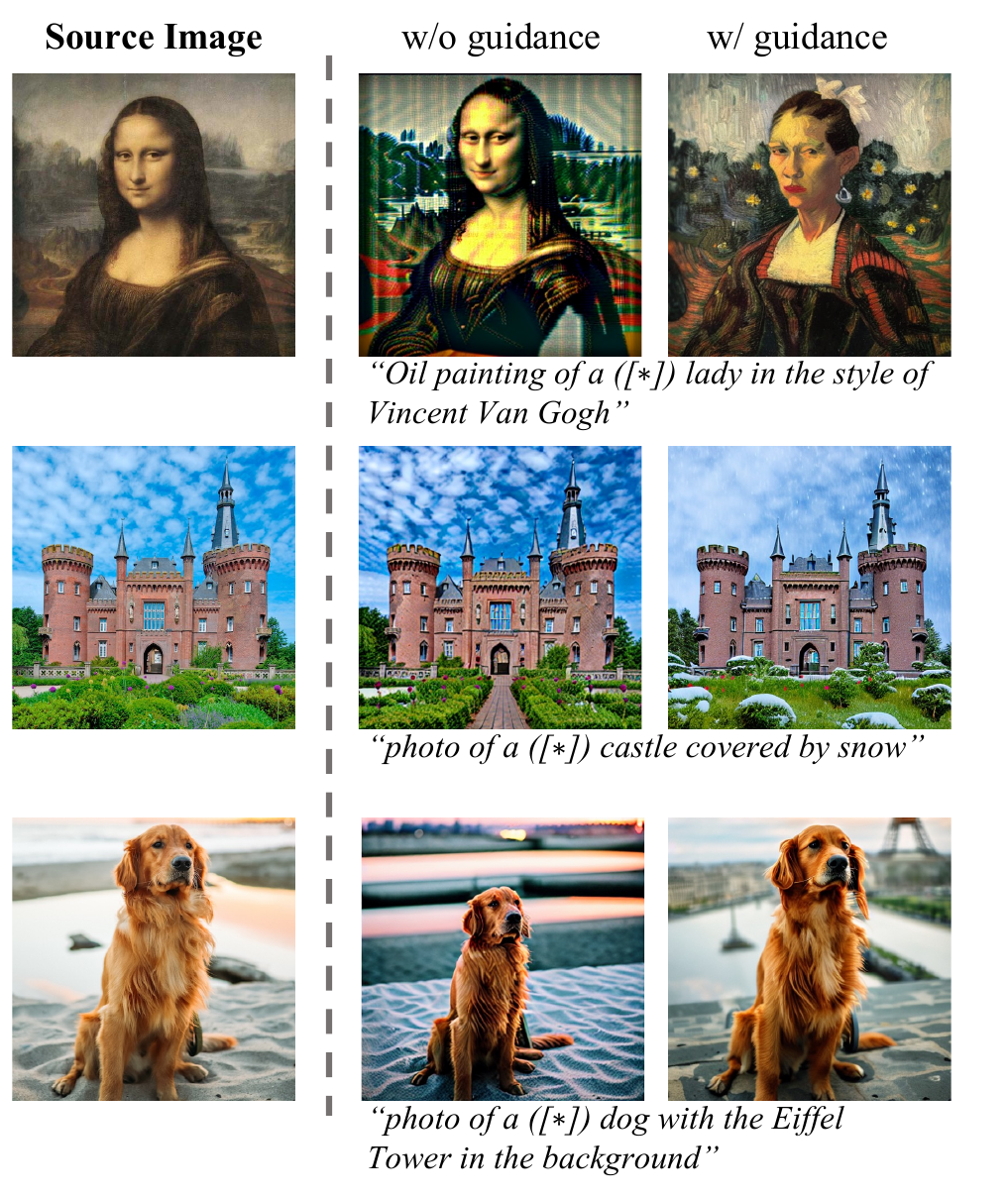}
    \caption{\textbf{Analysis of model-based classifier-free guidance.} 
    Directly sampling with target text using the fine-tuned model (w/o guidance) fails to generate images corresponding to the text prompt. In contrast, the model-based classifier-free guidance (w/ guidance) can synthesize high-fidelity images. }
    \label{fig: guidance_ablation}
\end{figure}

\noindent \textbf{Analysis of model-based classifier-free guidance.}
We generate editing results by directly sampling from the fine-tuned model using Eqn.~\ref{equ: classifier-free}.
We fine-tune the model without the patch-based schedule for $800$ steps to retain more generalization ability.
In this case, the model can perfectly reconstruct the source image while preserving as much editing ability as possible.
We denote the setting as \emph{w/o gudiance}. 
Using the same fine-tuned model, we conduct experiments under model-based classifier-free guidance, which we denote as \emph{w/ guidance}. 
As shown in Fig.~\ref{fig: guidance_ablation}, sampling without our model-based classifier-free guidance fails to react to the prompt, while our method can successfully edit images to match the target language guidance. We further analyze two hyper-parameters ($K$ and $v$) in model-based classifier-free guidance.

\begin{figure}
    \centering
    \includegraphics[width=\linewidth]{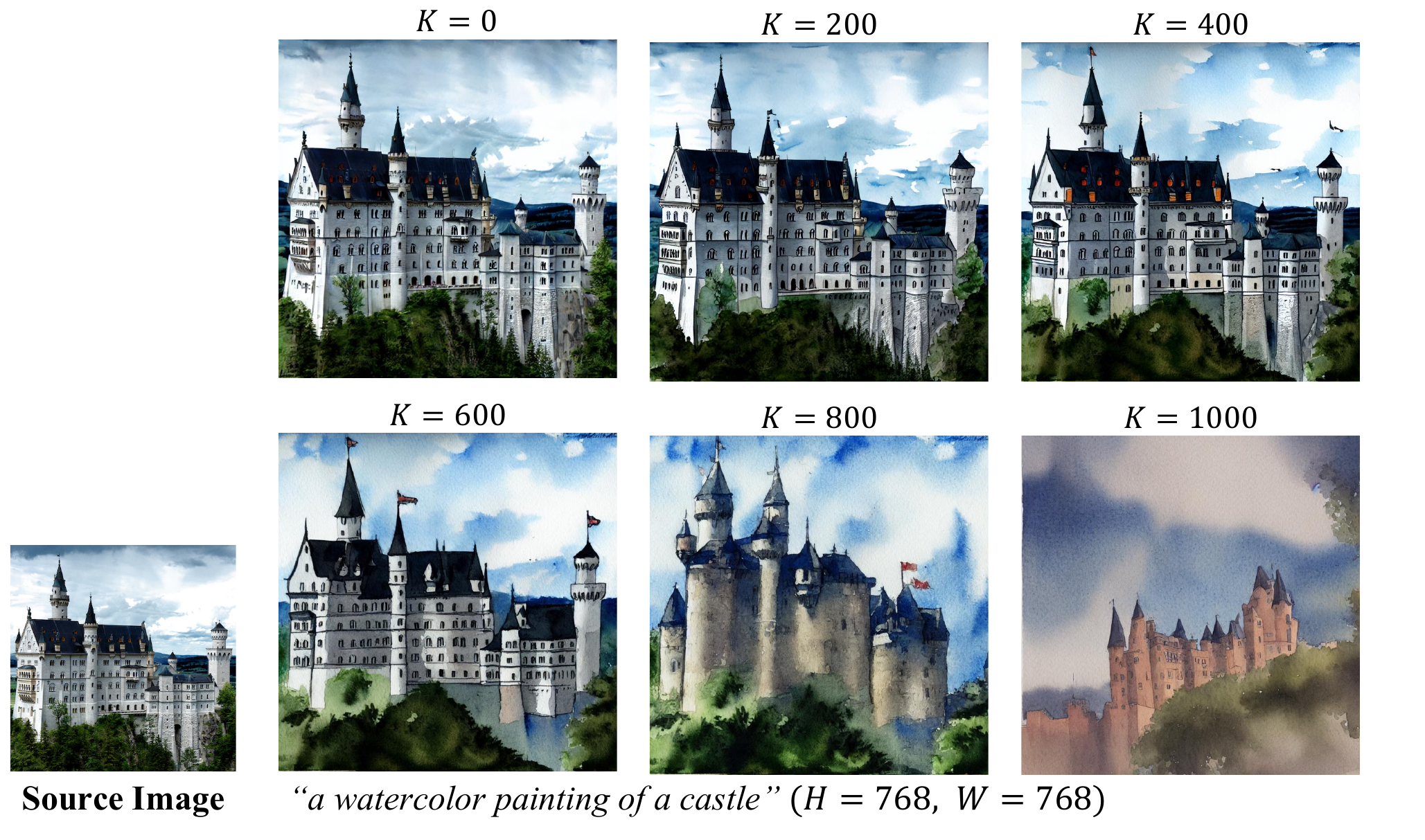}
    \caption{\textbf{Analysis on guidance step $K$.} Varying $K$, we can decide the steps where the model-based guidance is applied, which controls the details from the source image and edits to be applied.}
    \label{fig: ablation_K}
\end{figure}

\noindent \textbf{Analysis on guidance step $K$} in Sec.~\ref{sec: methods: guidance}.
In Fig.~\ref{fig: ablation_K}, we show our results on editing with different settings of $K$.
We conduct this set of experiments by editing one single image with the same language guidance at  $768\times768$ resolution.
We set $v=0.7$.
When $K=0$, the model-based classifier-free guidance is applied for each step of the denoising process.
Since the generalization ability of the fine-tuned model is limited, in this case, the model fails to apply the desired property to the single source image.
When $K=1,000$, the model-based classifier-free guidance is not applied to any step.
Thus, the structure of the image is not preserved, and the generated result becomes a random sample of the pre-trained model.

\begin{figure}[t]
  \centering
  \begin{subfigure}{0.49\linewidth}
    \centering
        \includegraphics[width=1\linewidth]{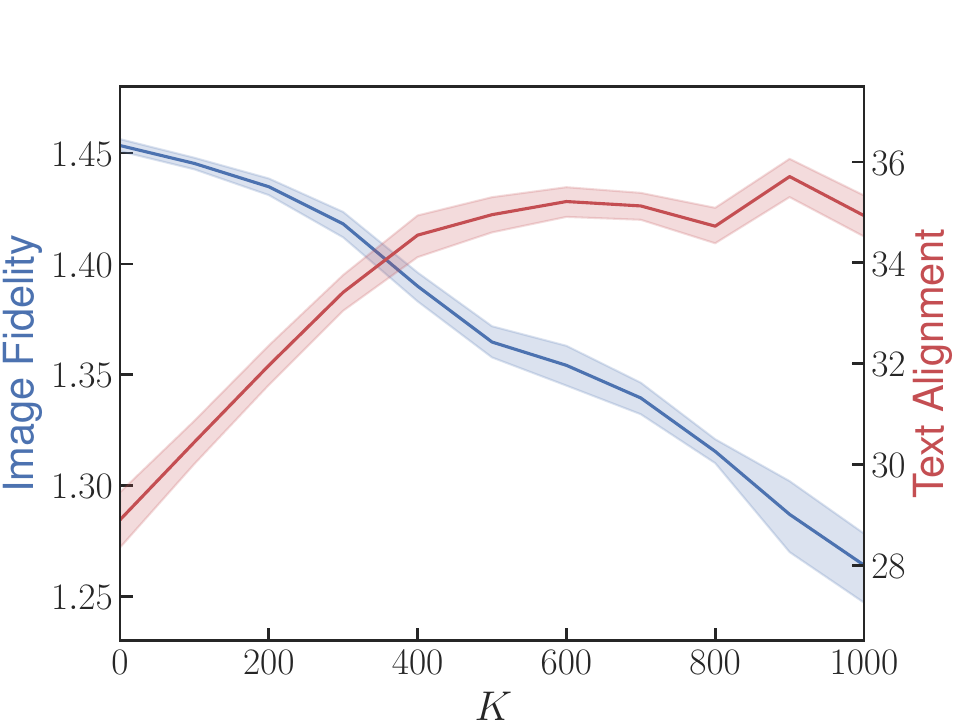}
    \caption{Number of steps of the guidance.}
    \label{fig: ablation_K_scores}
  \end{subfigure}
  \begin{subfigure}{0.49\linewidth}
    \centering
        \includegraphics[width=1\linewidth]{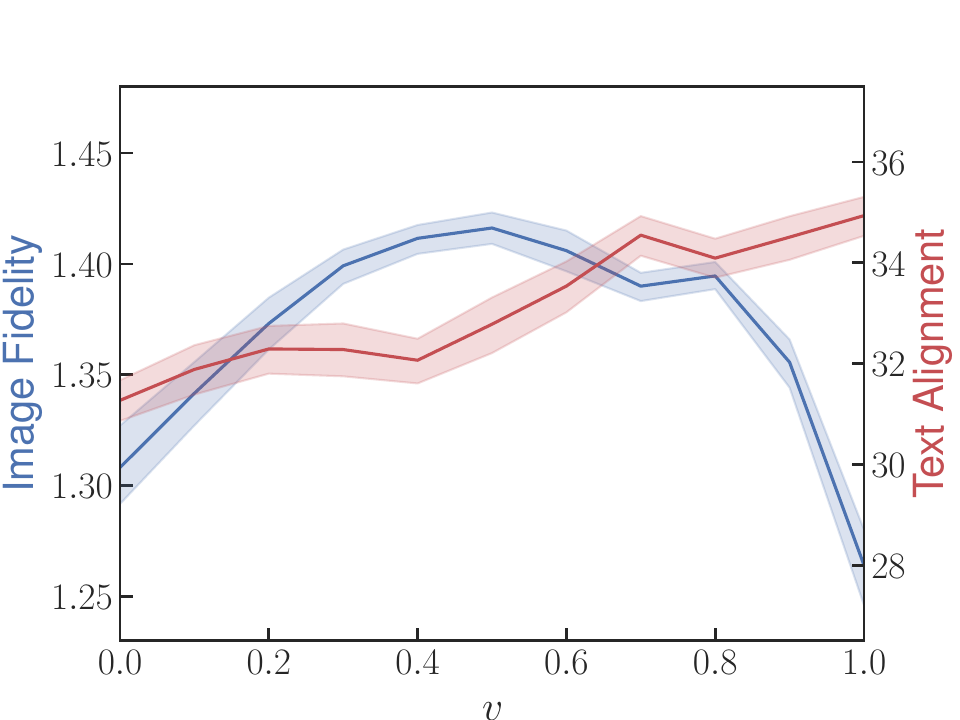}
    \caption{Weight of the guidance.}
    \label{fig:ablation_v_scores}
  \end{subfigure}
  \caption{
  \textbf{Trade-off between fidelity and alignment with target text.}
  We calculate the CLIP score~\cite{radford2021learning} (a higher CLIP score indicates a better alignment between the edit result and target text) and LPIPS score~\cite{zhang2018unreasonable} (showing the fidelity between edit results and source image, the higher, the better).
  }
  \label{fig:cost_comp}
\end{figure}

We further show the quantitative results Fig.~\ref{fig: ablation_K_scores}.
We repeat the abovementioned procedure over different $K$ and randomly sample $20$ images for each $K$.
We calculate two metrics. To understand the editing result, the \emph{image fidelity} that is measured 
by the LPIPS~\cite{zhang2018unreasonable} distance between the original and the edited image.
The \emph{text alignment} calculated by the CLIP~\cite{radford2021learning} score to understand the alignment between our generated images and target text.
As can be seen, the image fidelity drops with the increase of $K$, indicating more details provided by the pre-trained model instead of the fine-tuned one.
The text alignment measurement improves since the more details generated by the pre-trained model, the better editing results align with the target domain.
To preserve the edit result's authenticity and fidelity to the source image, we set $K$ as $400$.

\begin{figure}
    \centering    
    \includegraphics[width=\linewidth]{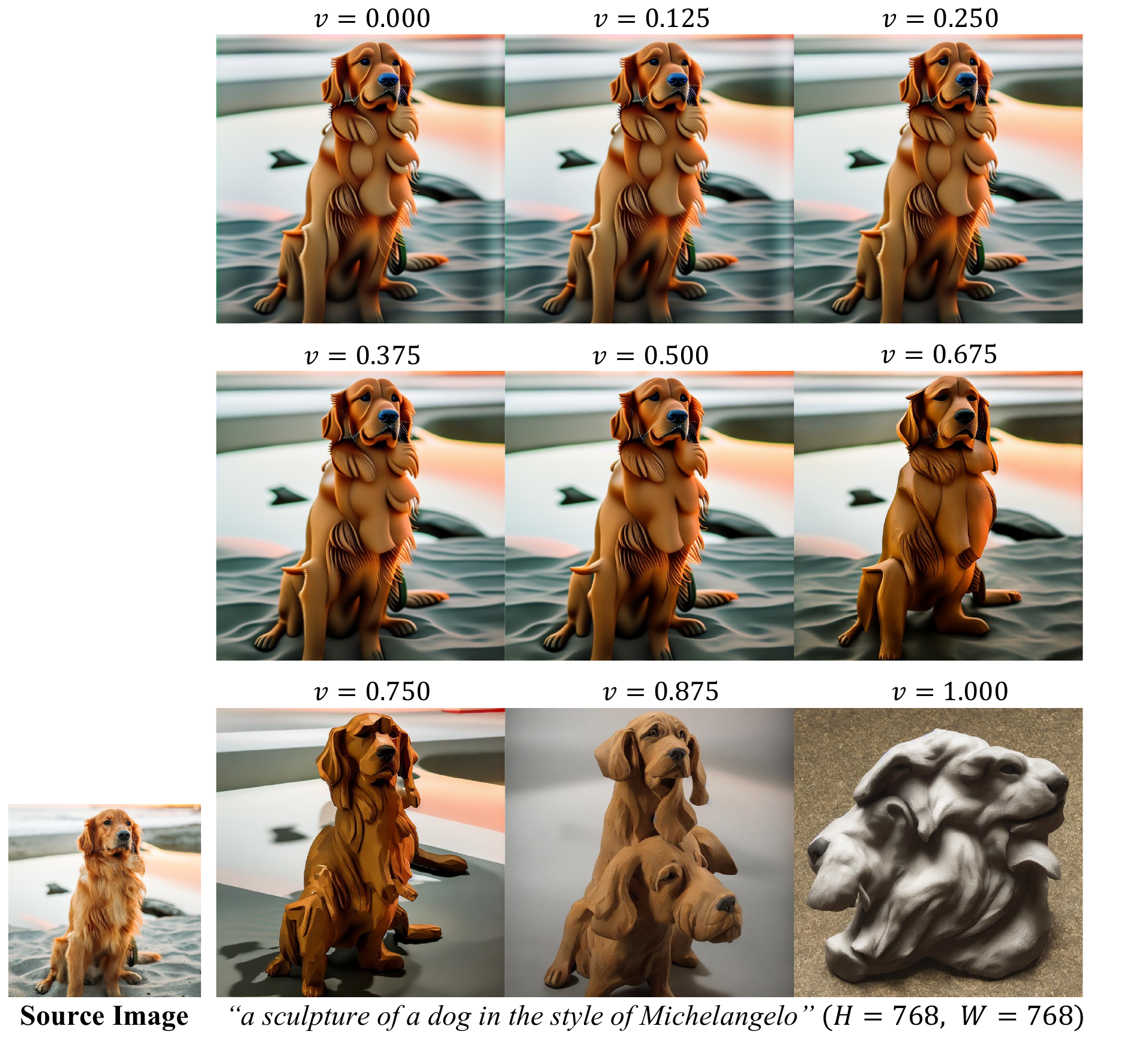}
    \caption{\textbf{Analysis of score interpolation.} Varying $v$ with the same random seed gets editing results with various qualities.}
    \label{fig: ablation_v}
\end{figure}

\noindent \textbf{Analysis on guidance weight $v$} in Sec.~\ref{sec: methods: guidance}.
We further study the impact of the guidance weight ($v$) in Fig.~\ref{fig: ablation_v}.
We set $K=400$ and resolution as $768\times768$ for each edit and use the same random seed to generate the result.
As can be observed, the value of $v$ controls the fidelity of the edit result.
However, since the pre-trained model is trained at the resolution $512\times512$, the generated image contains many artifacts.
When $v=1$, the synthesized image entirely depends on the results from the pre-trained model.
Additionally, we conduct quantitative experiments with LPIPS score measuring the image fidelity and CLIP score for the text alignment in Fig.~\ref{fig:ablation_v_scores}.
When $v$ is close to $1$, the fidelity decreases while the edited feature decreases.
When $v$ is close to $0$, the model relies mainly on the fine-tuned model for the output when $t> K$.
However, since the fine-tuned model contains poor generalization ability, there is a significant amount of artifacts in the generated results, which leads to a poor LPIPS score.
We choose $0.7$ for each edit in this work as a trade-off between fidelity and creativity.

\begin{figure*}
\setlength{\linewidth}{\textwidth}
\setlength{\hsize}{\textwidth}
\centering
\includegraphics[width=\textwidth]{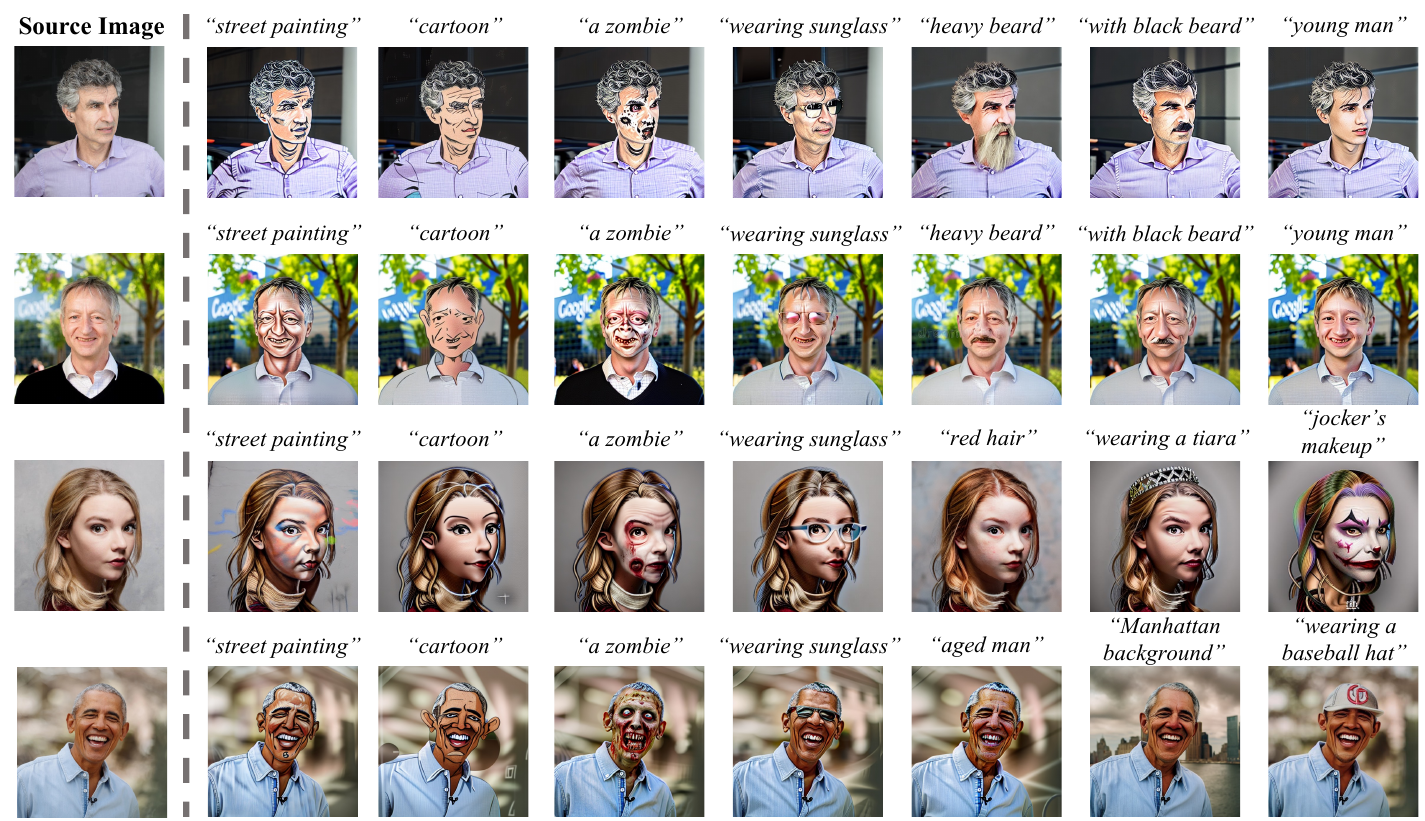}
\caption{\textbf{In-the-wild human face manipulation.} We conduct various editing on human face photos, locally or globally. The models are trained and edited at a resolution of $512\times512$. }
\label{fig:face edit}
\end{figure*}
\begin{figure*}
    \centering
    \begin{subfigure}{0.33\linewidth}
        \centering
        \includegraphics[width=1\linewidth]{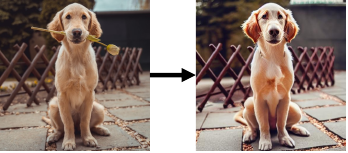}
        \caption{Content Removal}
        \label{fig:other applications:content removal}
    \end{subfigure}
    \begin{subfigure}{0.33\linewidth}
        \centering
        \includegraphics[width=1\linewidth]{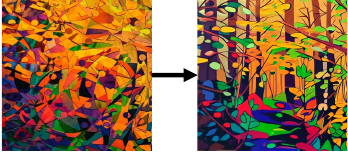}
        \caption{Style Generation}
        \label{fig:other applications:style generation}
    \end{subfigure}
    \begin{subfigure}{0.33\linewidth}
        \centering
        \includegraphics[width=1\linewidth]{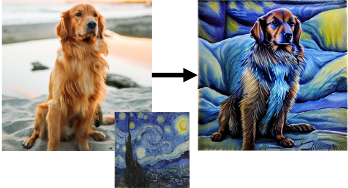}
        \caption{Style Transfer}
        \label{fig:other applications:style transfer}
    \end{subfigure}
    \caption{\textbf{More applications.} We show how our approach can be applied to various tasks in image editing, such as content removal (a), style generation (b), and style transfer (c).}
    \label{fig:other applications}
\end{figure*}

\subsection{More Editing Tasks}

\noindent\textbf{Face manipulation.}
Our method demonstrates promising editing ability for \emph{in-the-wild human faces}.
As shown in \cref{fig:face edit}, our approach can edit locally and globally on human faces for various facial manipulation tasks, \emph{e.g.}, image stylization, adding accessories, and age changing.

\noindent\textbf{Content removal.} In \cref{fig:other applications:content removal}, we show the content removal using our method. We fine-tune the pre-trained large-scale text-to-image model with the language descriptor as ``a [$\ast$] dog with a flower in mouth''.
For sampling, we use text prompts such as ``a dog'' and ``a [$\ast$] dog'' for the pre-trained and fine-tuned models.
The pre-trained model successfully removes the flower held in the mouth of the dog.

\noindent\textbf{Style generation.} Our method can also be employed to learn the underlying style of an image. As shown in \cref{fig:other applications:style generation}, the model is fine-tuned with the text, ``a  painting in the [$\ast$] style''.
When sampling results, we feed the pre-trained model a prompt as ``painting of a forest''. 
The model can successfully synthesize images with the specified content in the style of the given real image.

\noindent\textbf{Style transfer.} Our model-based classifier-free guidance can be leveraged to combine multiple models for providing the guidance.
We show the result in \cref{fig:other applications:style transfer} by doing a style transfer task with dual-model guidance.
We fine-tune two models using prompts: ``picture of a [$\ast$] dog'' and ``painting in [$\ast$] style''.
During inference, we give the pre-trained model the prompt ``painting of a dog'' and fine-tuned models with prompts the same as training.
With guidance from two separate models, our method can generate images with the content from one and style from the other and achieve stylized generation.

\section{Conclusion}

This work introduces SINE, a method for single-image editing. 
With only one image and a brief description of the object in the image, our approach can enable a wide range of editing for arbitrary resolution, followed by the information depicted in the language guidance.
To achieve such results, we leverage the pre-trained large-scale text-to-image diffusion model.
Specifically, we first fine-tune the pre-trained model with our patch-based fine-tuning method until it overfits the single image.
Then, during sampling time, we use the overfitted model to guide the pre-trained diffusion model for image synthesis, which maintains the fidelity of the results while taking advantage of the generalization ability of the pre-trained model.
Compared with other methods, our approach has a better geometrical understanding of the image and thus can conduct complex editing to the images besides style transfer.

However, in some cases where confusing editing guidance is given for the diffusion model, \eg, a chair-shaped dog, our method could fail.
In cases where drastic changes are to be applied, \eg, changing a dog to a tiger in the same posture, there are also noticeable artifacts.
We show more examples in the Appendix.

One future direction is improving the fidelity of the editing results, which could be achieved by alleviating the overfitting problem of the fine-tuned model.

\noindent {\bf Acknowledgments.} 
This research has been partially funded by research grants to D. Metaxas through NSF IUCRC CARTA-1747778, 2235405, 2212301, 1951890, 2003874, 2310966, FA9550-23-1-0417, and NIH-5R01HL127661.

{\small
\bibliographystyle{ieee_fullname}
\bibliography{egbib}
}
\clearpage
\appendix

\section*{Appendix}
In the Appendix, we provide the following:
\setlist{nosep}
\begin{itemize}
    \item More comparisons with exiting works on editing single real image (\cref{supp: sec: comparison}).
    \item More results for applying SINE on editing a single image and the novel image manipulation tasks that can be enabled by our approach (\cref{supp: sec: additional results}).
    \item More ablation analysis (\cref{supp: sec: ablations}).
    \item Discussion about the limitation of our method and possible future work (\cref{supp: sec: limitations}).
\end{itemize}
\section{More Comparisons}
\label{supp: sec: comparison}
Besides the comparison with existing works shown in the main paper, we provide more results by comparing our approach with 
Prompt-to-Promt~\cite{hertz2022prompt}.
In addition, we compare our methods with training-free single-image editing approaches, including SDEdit~\cite{meng2021sdedit} and ILVR~\cite{choi2021ilvr}.

We first show the technical differences between our works and training-free methods in \cref{tab:mbcfg}.
SDEdit~\cite{meng2021sdedit} applies the diffusion process on an image or a user-created semantic map to conduct the denoising procedure, conditioned with the desired output. 
ILVR~\cite{choi2021ilvr} guides the denoising process by replacing the low-frequency part of the sample with that of the target reference image.

\begin{figure}[h]
\centering
\includegraphics[width=\linewidth]{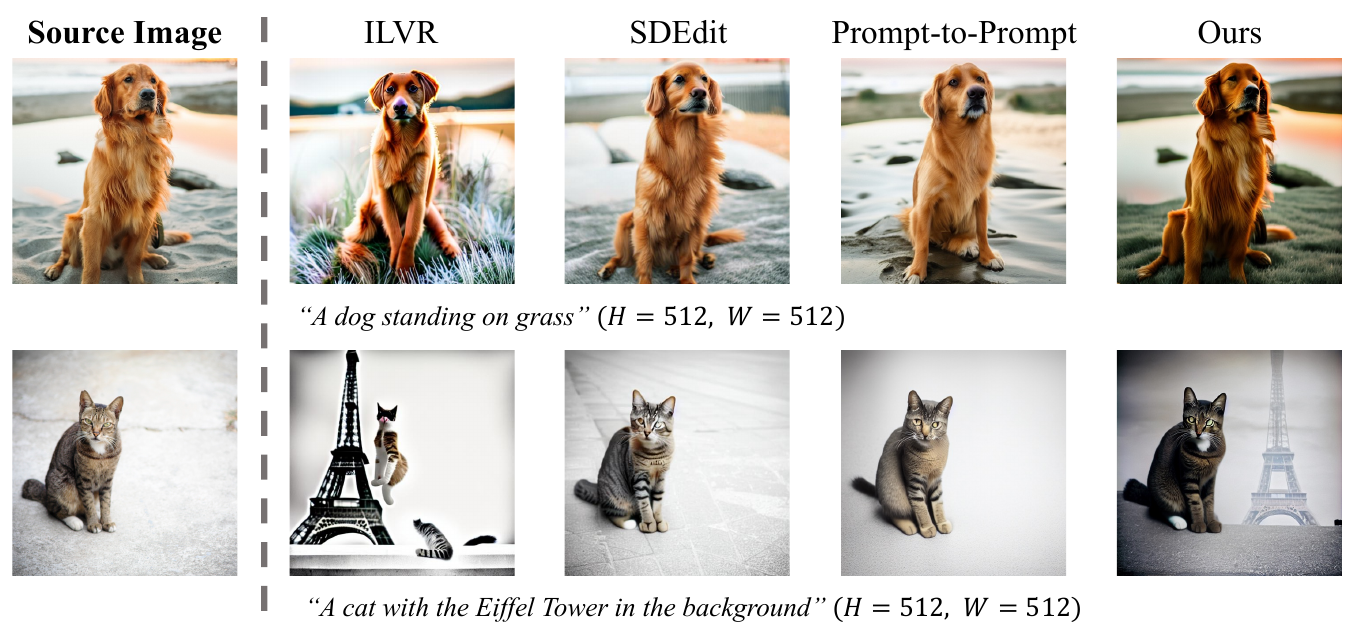}
\caption{\textbf{Comparison results.}
We compare our method with ILVR~\cite{choi2021ilvr}, SDEdit~\cite{meng2021sdedit}, and Prompt-to-Prompt~\cite{hertz2022prompt} on editing single real image. Note that when the hyper-parameter $N$ is set to $1$, the process of ILVR is equivalent to stochastic SDEdit. We adopt the official implementation of ILVR for conducting experiments on SDEdit. For ILVR, we set downsample ratio of $N=8$. For SDEdit, we use the scholastic q sample. In both cases, we set $K=400$.
}
\label{fig: comparasion}
\end{figure}

The visual comparisons are illustrated in \cref{fig: comparasion}. As can be seen, our approach significantly outperforms other methods for generating high-fidelity images with the maximal keeping of the details in the source image.

\begin{table*}[h]
\caption{The differences between our approach and other training-free methods for single image editing.}\label{tab:mbcfg}
\centering
\begin{tabular}{lcccc}
\toprule
Guidance    & Finetune & Compatible w/ LDM~\cite{rombach2022high} & Position Control & Admits Multiple Inputs \\\hline
SINE (Ours) & \textcolor{myred}{Required}       & \greencmark & \greencmark & \greencmark \\
ILVR~\cite{choi2021ilvr} & \textcolor{mygreen}{Not Required} & \redxmark   & \redxmark   & \redxmark   \\
SDEdit~\cite{meng2021sdedit} & \textcolor{mygreen}{Not Required} & \greencmark & \redxmark   & \redxmark   \\
\bottomrule
\end{tabular}
\end{table*}

\section{More Editing Results}
\label{supp: sec: additional results}
\begin{figure*}
\setlength{\linewidth}{\textwidth}
\setlength{\hsize}{\textwidth}
\centering
\includegraphics[width=\textwidth]{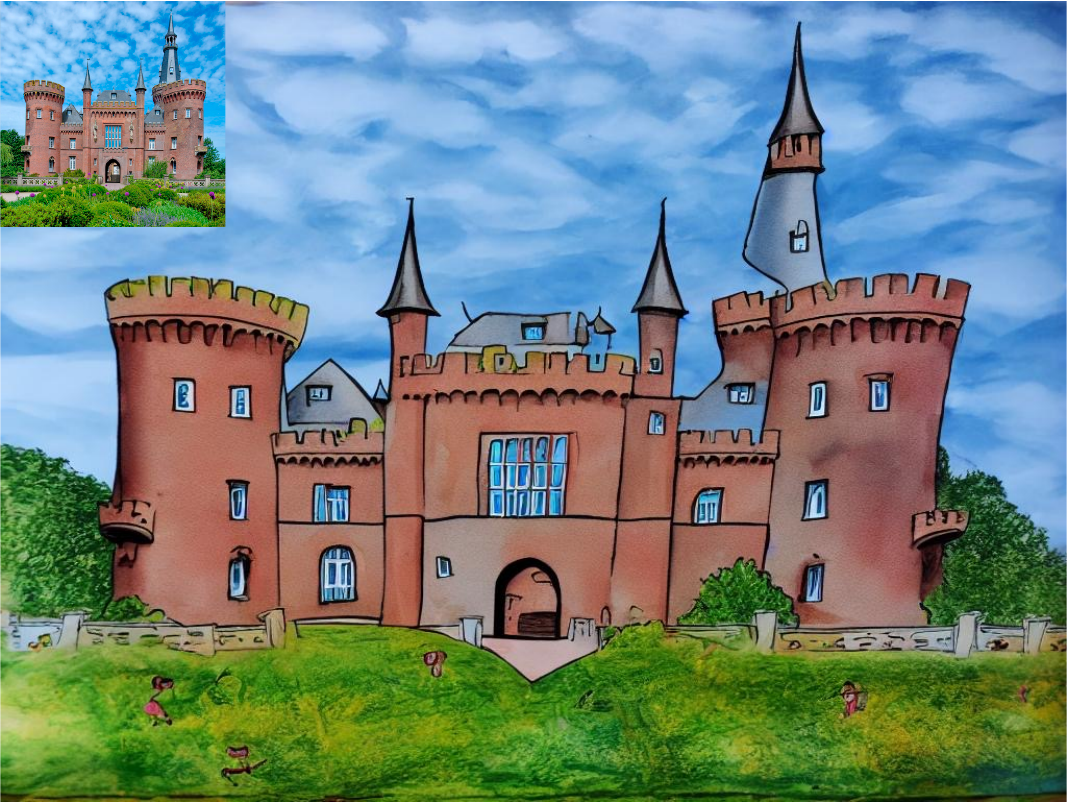}
\caption{\textbf{A children's painting of a castle.}
The generation resolution is set to $H=768~\text{and}~W=1024$.
We use $K=400$ and $v=0.7$ in this sample.
}
\label{fig:high res: castle_children}
\end{figure*}

\begin{figure*}
\setlength{\linewidth}{\textwidth}
\setlength{\hsize}{\textwidth}
\centering
\includegraphics[width=\textwidth]{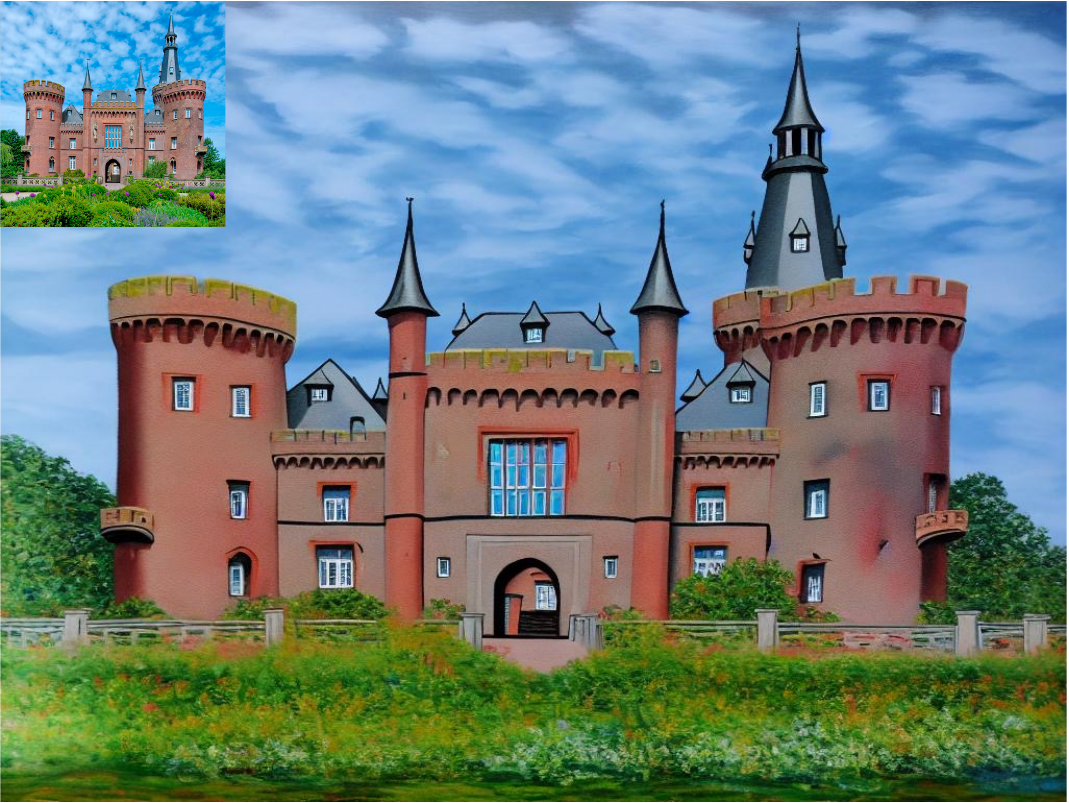}
\caption{\textbf{A painting of a castle in the style of Claude Monet.}
The output resolution is set to $H=768~\text{and}~W=1024$.
We use $K=400$ and $v=0.65$ in this example.}
\label{fig:high res: castle_monet}
\end{figure*}

\begin{figure*}
\setlength{\linewidth}{\textwidth}
\setlength{\hsize}{\textwidth}
\centering
\includegraphics[width=\textwidth]{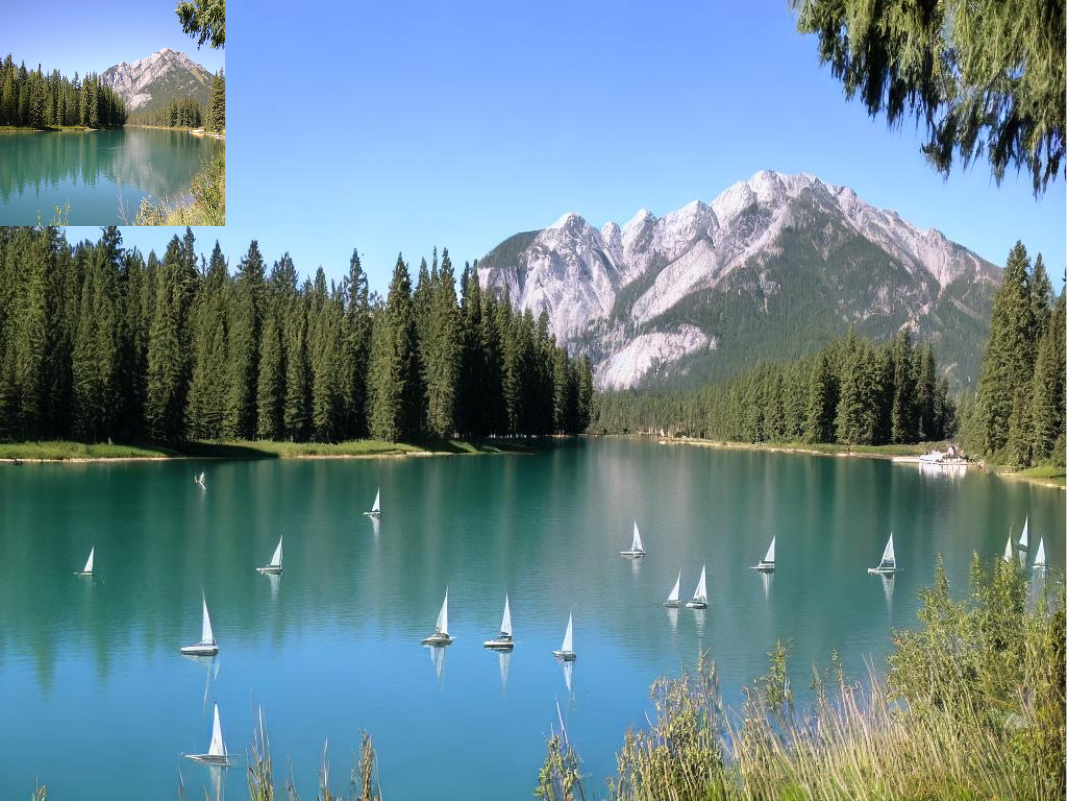}
\caption{\textbf{A photo of a lake with many sailboats.}  
The output resolution is set to $H=768~\text{and}~W=1024$.
We use $K=400$ and $v=0.7$ in this case.}
\label{fig:high res: lake sailboats}
\end{figure*}

\begin{figure*}
\setlength{\linewidth}{\textwidth}
\setlength{\hsize}{\textwidth}
\centering
\includegraphics[width=\textwidth]{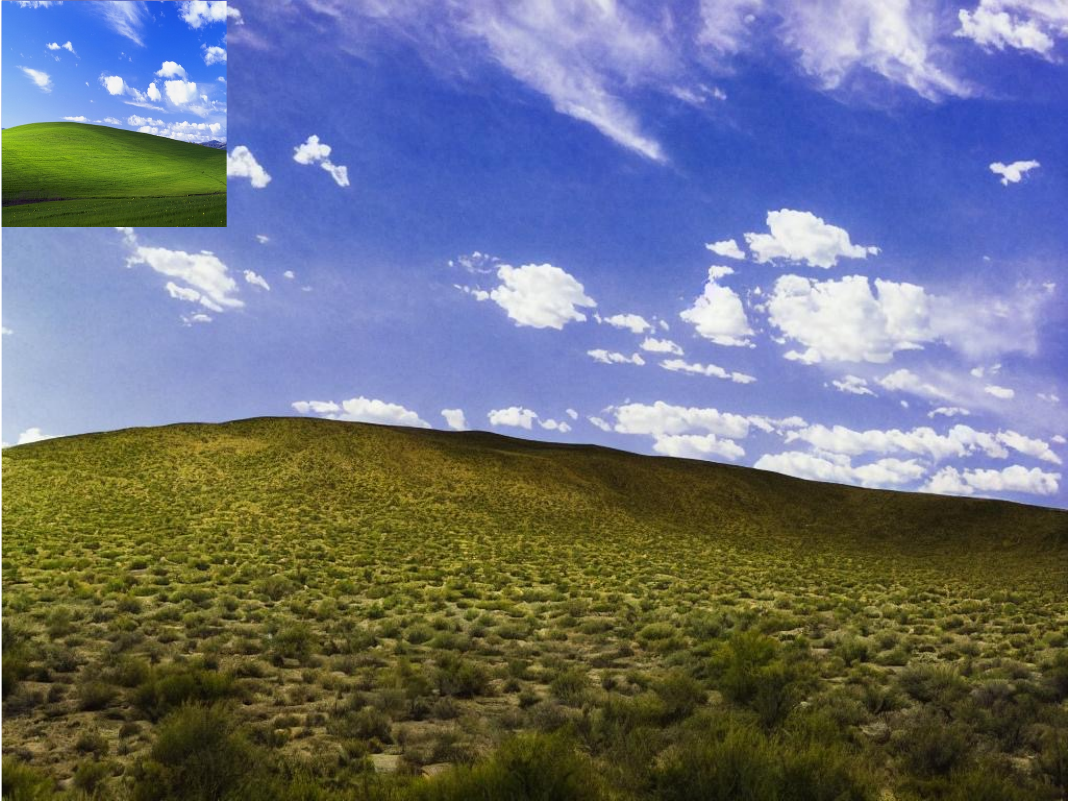}
\caption{\textbf{A desert.} The output resolution is set to $H=768~\text{and}~W=1024$.
We use $K=500$ and $v=0.8$ in this case. }
\label{fig:high res: desert}
\end{figure*}

\begin{figure*}
\setlength{\linewidth}{\textwidth}
\setlength{\hsize}{\textwidth}
\centering
\includegraphics[width=\textwidth]{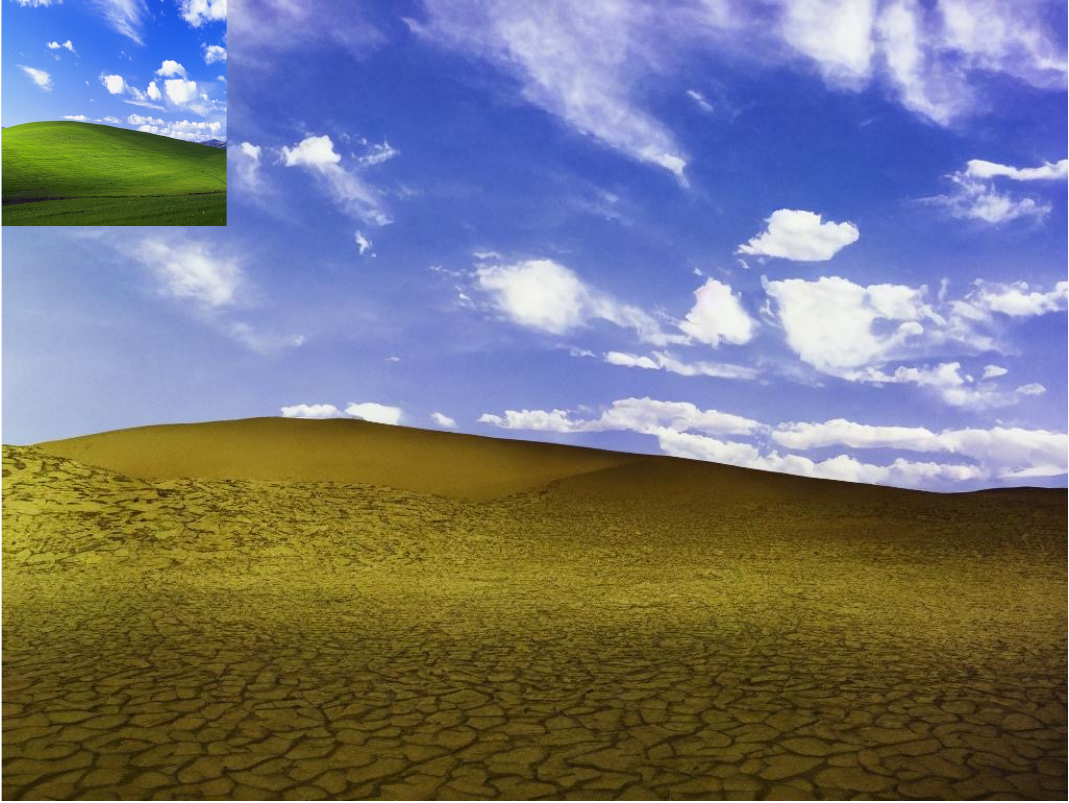}
\caption{\textbf{A desert.} The output resolution is set to $H=768~\text{and}~W=1024$.
We use $K=500$ and $v=0.8$ in this case.  }
\label{fig:high res: desert1}
\end{figure*}

\begin{figure*}
\setlength{\linewidth}{0.8\textwidth}
\setlength{\hsize}{\textwidth}
\centering
\includegraphics[width=0.8\textwidth]{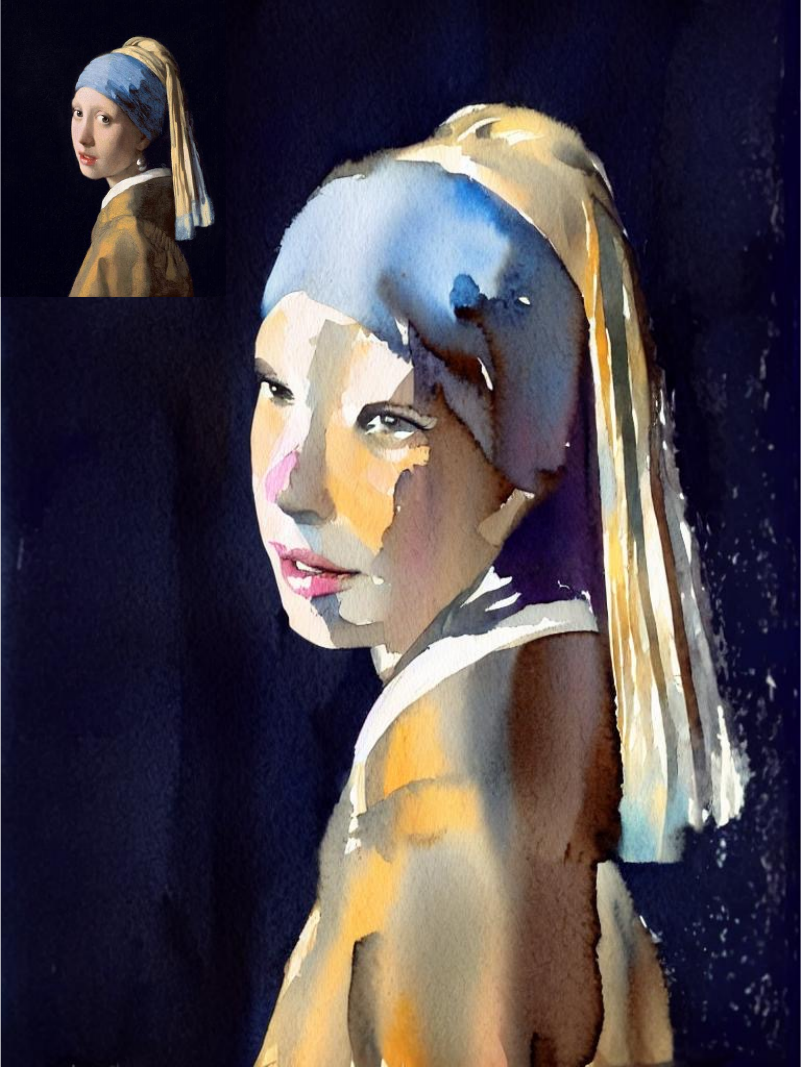}
\caption{\textbf{A watercolor painting of a girl.}  The output resolution is set to $H=1024~\text{and}~W=768$.
We use $K=400$ and $v=0.6$ in this case.}
\label{fig:high res: girl watercolor}
\end{figure*}

We provide more editing results in \cref{fig:high res: castle_children}, \cref{fig:high res: castle_monet}, \cref{fig:high res: lake sailboats}, \cref{fig:high res: desert}, \cref{fig:high res: desert1}, and \cref{fig:high res: girl watercolor}.
All results are obtained by fine-tuning the large-scale text-to-image model~\cite{rombach2022high} using our proposed patch-based method at the resolution of $512\times512$ and sampling with our introduced model-based classifier-free guidance at a higher resolution, \emph{e.g.}, $768\times1024$.
Images on the top-left corner of these results are the real images utilized for fine-tuning.
We specify the hyper-parameters used during sampling in the caption of each image.

\section{More Ablations}
\label{supp: sec: ablations}

\begin{figure*}[h]
\setlength{\linewidth}{\textwidth}
\setlength{\hsize}{\textwidth}
\centering
    \begin{subfigure}{0.33\linewidth}
    \centering
        \includegraphics[width=0.9\linewidth]{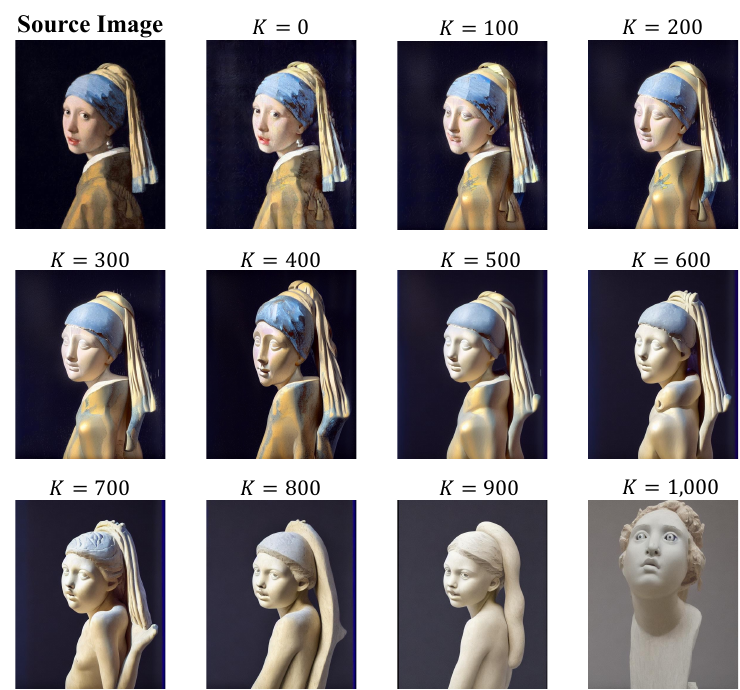}
        \caption{We set $v=0.5$ and change $K$}
    \end{subfigure}
    \begin{subfigure}{0.33\linewidth}
    \centering
        \includegraphics[width=0.9\linewidth]{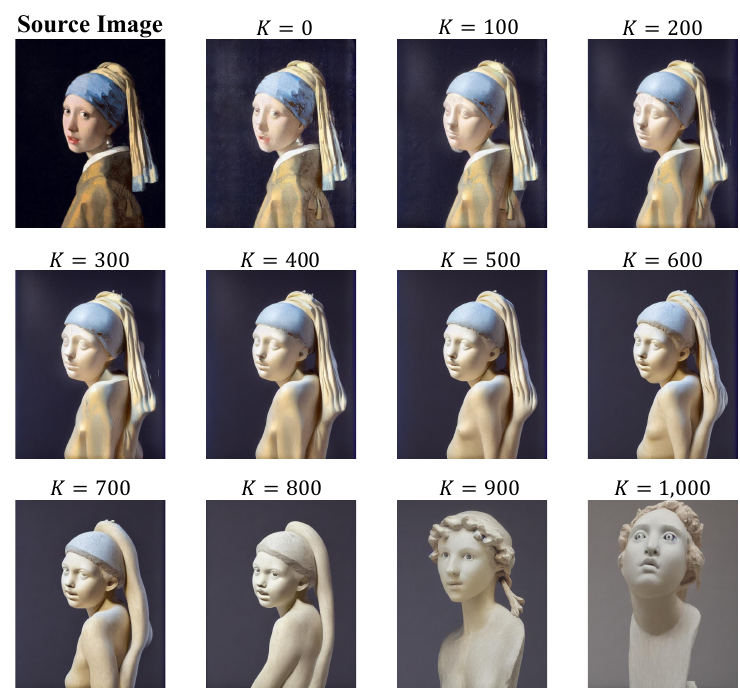}
        \caption{We set $v=0.7$ and change $K$.}
    \end{subfigure}
    \begin{subfigure}{0.33\linewidth}
    \centering
        \includegraphics[width=0.9\linewidth]{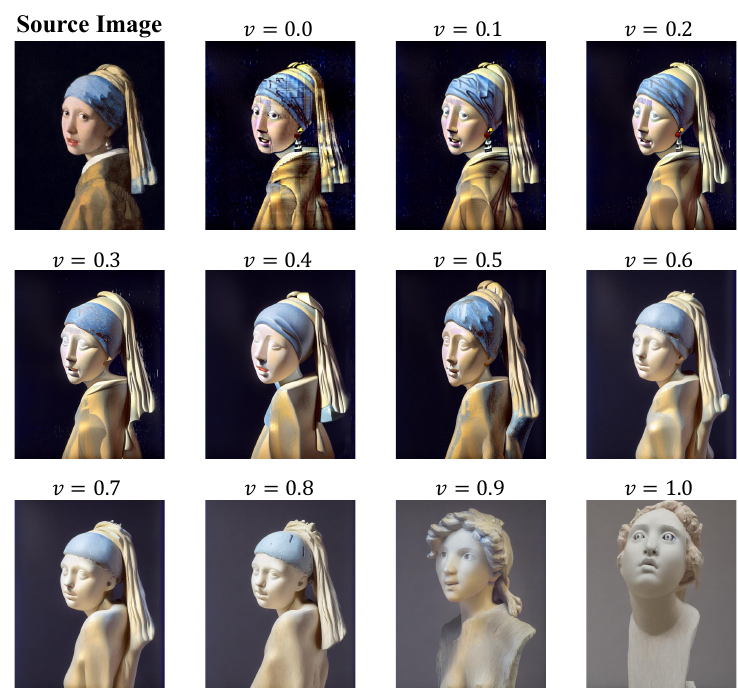}
        \caption{We set $K=400$ and change $v$}
    \end{subfigure}
    
    \caption{\textbf{``A sculpture of a girl''} with the resolution of $H=640~\text{and}~W=512$.}
    \label{fig: ablation: girl}
\end{figure*}

\begin{figure*}
\setlength{\linewidth}{\textwidth}
\setlength{\hsize}{\textwidth}
\centering
    \begin{subfigure}{0.33\linewidth}
    \centering
        \includegraphics[width=0.9\linewidth]{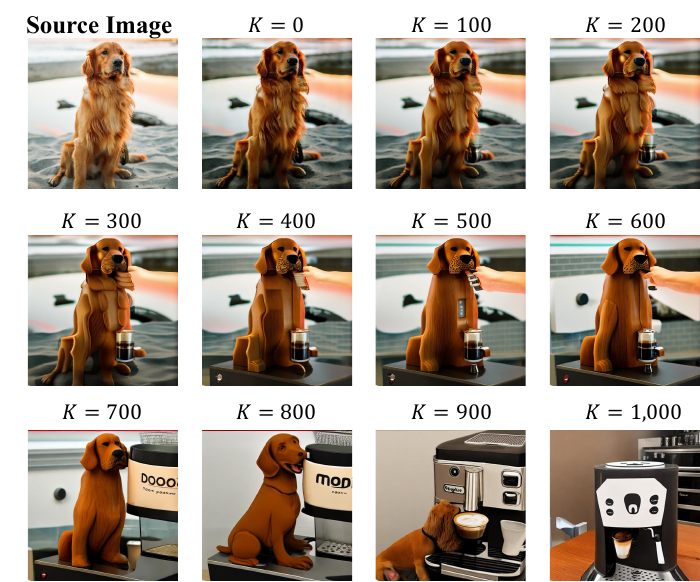}
        \caption{We set $v=0.7$ and change $K$}
    \end{subfigure}
    \begin{subfigure}{0.33\linewidth}
    \centering
        \includegraphics[width=0.9\linewidth]{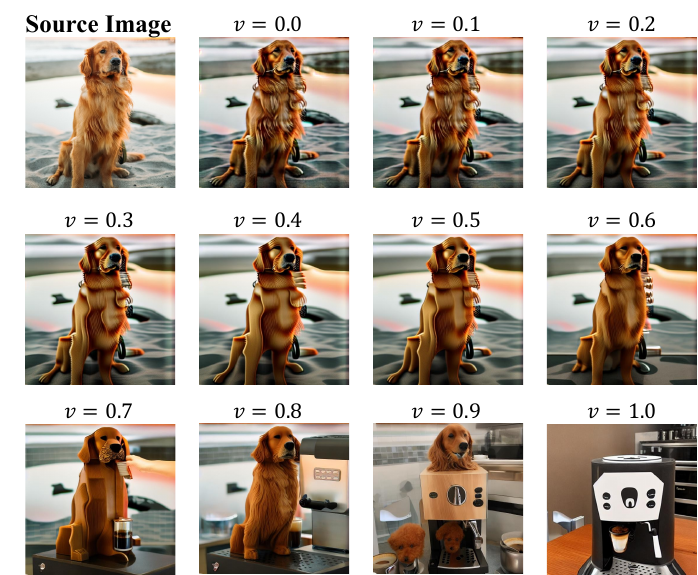}
        \caption{We set $K=400$ and change $v$}
    \end{subfigure}
    \begin{subfigure}{0.33\linewidth}
    \centering
        \includegraphics[width=0.9\linewidth]{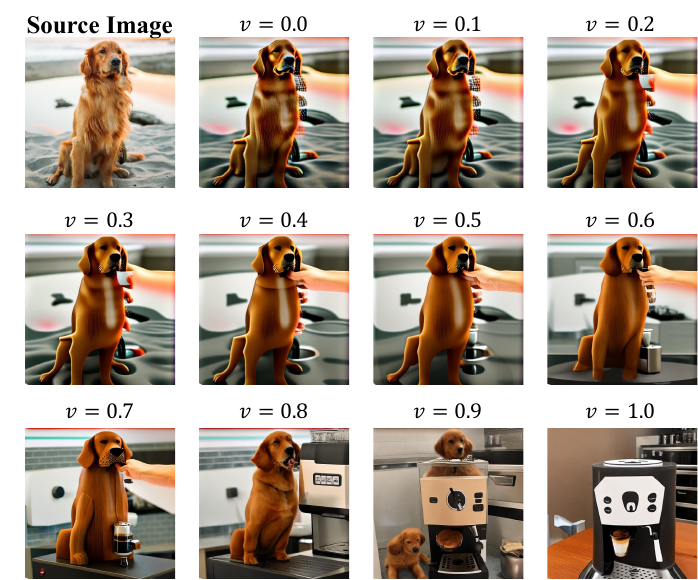}
        \caption{We set $K=600$ and change $v$}
    \end{subfigure}
    
    \caption{\textbf{``A coffee machine in the shape of a dog''} with the resolution of $H=512~\text{and}~W=512$.}
    \label{fig: ablation: dog}
\end{figure*}

\begin{figure*}
\setlength{\linewidth}{\textwidth}
\setlength{\hsize}{\textwidth}
\centering
    \begin{subfigure}{0.48\linewidth}
    \centering
        \includegraphics[width=0.9\linewidth]{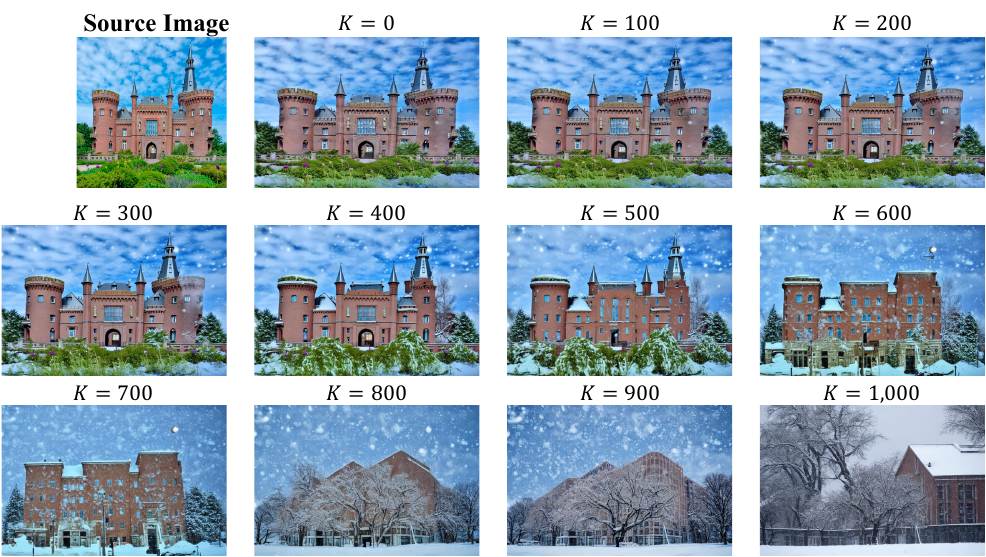}
        \caption{Set $v=0.7$ varying $K$}
    \end{subfigure}
    \begin{subfigure}{0.48\linewidth}
    \centering
        \includegraphics[width=0.9\linewidth]{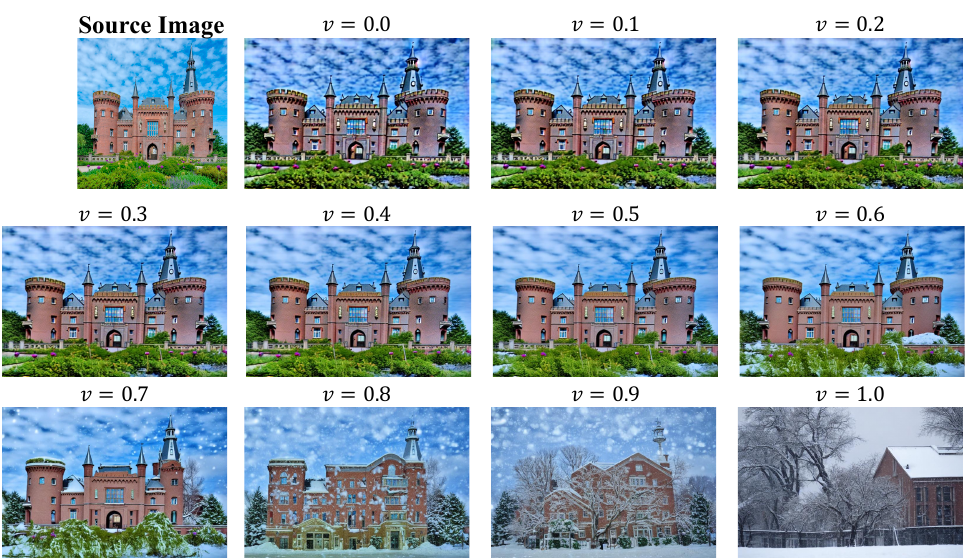}
        \caption{Set $K=400$ varying $v$}
    \end{subfigure}
    
    \caption{\textbf{``A castle covered by snow''} with the resolution of $H=512~\text{and}~W=768$.}
    \label{fig: ablation: castle}
\end{figure*}

\noindent \textbf{Analysis on guidance step $K$ and guidance weight $v$}.
We conduct experiments by varying the guidance step $K$ and guidance weight $v$ in \cref{fig: ablation: girl}, \cref{fig: ablation: dog}, and \cref{fig: ablation: castle}.
We use the same random seed and generate results with specific text prompts at a fixed resolution by varying the parameters.
These experiments show the same behavior of our approach as mentioned in Sec~\ref{sec: exp: ablation}.
By adjusting these two parameters, we can find an optimal combination specifically for the image and the target language guidance.
In most cases, we adopt the parameters setting of $K=400$ and $v=0.7$.
However, we want our model to maintain more fidelity or apply a stronger edit in some instances.
For example, the ``optimal'' setting we decide for experiments in \cref{fig: ablation: girl} is $v=0.5$ and $K=400$.

\begin{figure}
  \centering
  \begin{minipage}[c]{1\linewidth}
          \includegraphics[width=\linewidth]{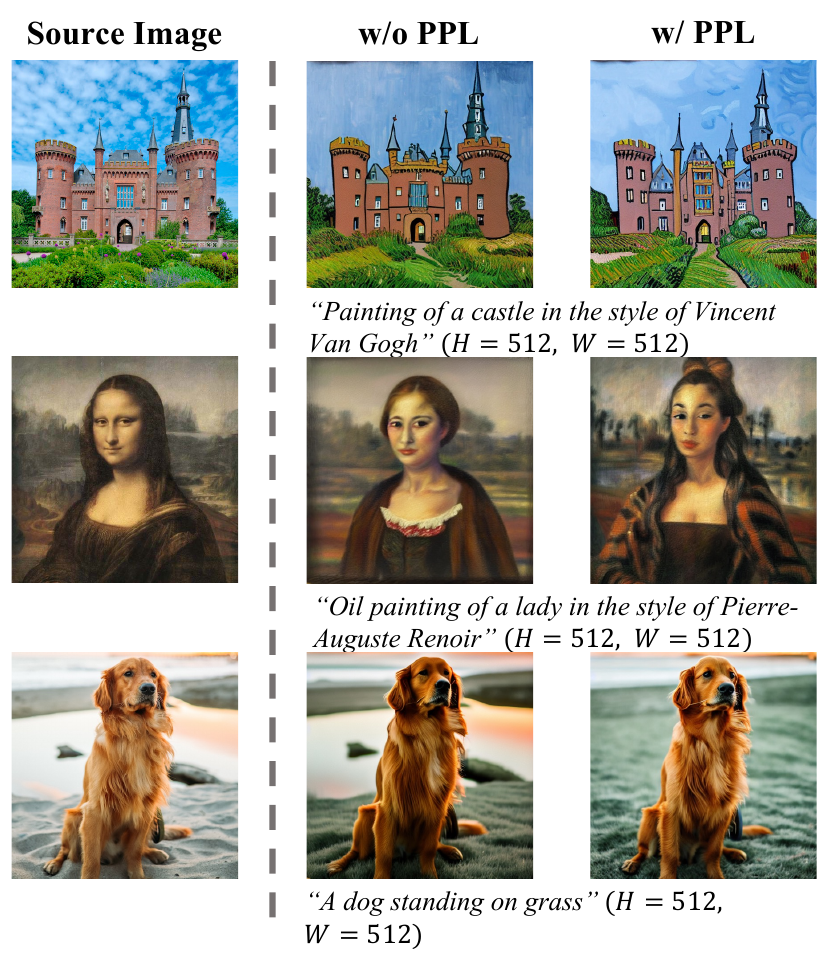}
          \caption{\textbf{Analysis of Prior-Preservation Loss (PPL).}}
    \label{fig:ablation regularization}
  \end{minipage}
\\
\begin{minipage}[c]{1\linewidth}
          \includegraphics[width=\linewidth]{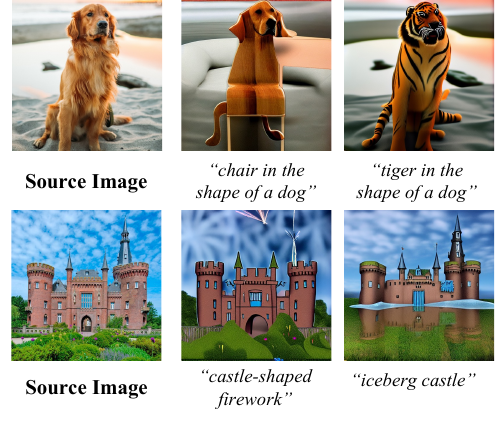}
    \caption{\textbf{Failure cases.} We showcase where our method fails to generate results with high fidelity and text alignment.}
    \label{fig:failure cases}
  \end{minipage}
\end{figure}

\noindent \textbf{Analysis on regularization loss}.
Dreambooth~\cite{ruiz2022dreambooth} proposes to leverage Prior-Preservation Loss(PPL) to address the issues of overfitting and language drift.
They propose to generate $200$ samples with the pre-trained model using the prompt ``a [class noun]''.
Then, during fine-tuning, they use these samples to regulate the model with the Prior-Preservation Loss to maintain the generalization ability of the model.
However, in our experiments, as shown in \cref{fig:ablation regularization}, this loss does not improve the final results due to the uniqueness of certain pictures/paintings.
On the contrary, more artifacts are introduced to the results, and the fidelity of the editing results decreases.
Therefore, given the motivation of editing unique images, we forfeit the generalization ability provided by regularizing the model with the samples generated by the pre-trained model.
We encourage our model to overfit a single image for the fidelity of the editing results.

\section{Limitations}
\label{supp: sec: limitations}

We present some failure cases in \cref{fig:failure cases}.
As mentioned in the main paper, when confusing guidance is given to the model or drastic change is to be applied, our method produces unsatisfying results.
The language comprehension limitation of the pre-trained model and the over-fitting issue of our fine-tuned model can cause this.
It would be an interesting future direction to explore how to over-fit on one single image without ``forgetting'' prior knowledge.

Also, as can be noticed in the second row of \cref{fig:face edit}, the color of the sweater is changed in most cases.
Also, the background letters are twisted after editing.
Even though our method can perform editing with maximal protection of the details in the source image, editing strictly on a specific part of an image is also worth further exploration.

\end{document}